\documentclass[10pt,twocolumn,letterpaper]{article}

\usepackage[pagenumbers]{cvpr}

\definecolor{cvprblue}{rgb}{0.21,0.49,0.74}
\usepackage[pagebackref,breaklinks,colorlinks,allcolors=cvprblue]{hyperref}
\usepackage{xcolor}
\usepackage{capt-of}
\usepackage{cuted}

\title{RenderFlow: Single-Step Neural Rendering via Flow Matching}

\author{
    Shenghao Zhang\textsuperscript{1,2} \quad
    Runtao Liu\textsuperscript{1} \quad
    Christopher Schroers\textsuperscript{1} \quad
    Yang Zhang\textsuperscript{1} \\[1em]
    \textsuperscript{1}Disney Research\textbar{}Studios \qquad
    \textsuperscript{2}ETH Z\"urich
}

\begin{document}
\maketitle

\begin{strip}
    \centering
    \includegraphics[width=1\textwidth,height=5.2cm,keepaspectratio]{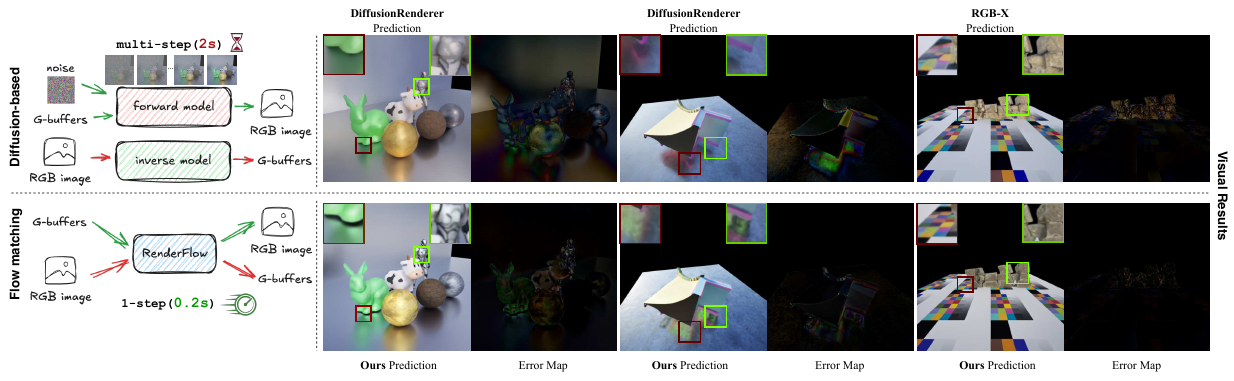}
    \vspace{-0.5em}
    \captionof{figure}{\textbf{Overview of the proposed \emph{RenderFlow} framework.} \textbf{Left:} \emph{RenderFlow} learns a single-step conditional flow from \emph{albedo} (rather than noise) to shaded images, yielding \textbf{$\sim$10$\times$ faster} rendering while better physical light transport; both forward and inverse rendering are unified within a \emph{single} model. \textbf{Right:} Example visualizations and corresponding error maps illustrating fidelity. } 
    \label{fig:teasers}
\vspace{-0.5em}
\end{strip}

\begin{abstract}
Conventional physically based rendering (PBR) pipelines generate photorealistic images through computationally intensive light transport simulations. Although recent deep learning approaches leverage diffusion model priors with geometry buffers (G-buffers) to produce visually compelling results without explicit scene geometry or light simulation, they remain constrained by two major limitations. First, the iterative nature of the diffusion process introduces substantial latency. Second, the inherent stochasticity of these generative models compromises physical accuracy and temporal consistency. In response to these challenges, we propose a novel, end-to-end, deterministic, single-step neural rendering framework, \textit{RenderFlow}, built upon a flow matching paradigm. To further strengthen both rendering quality and generalization, we propose an efficient and effective module for sparse keyframe guidance. Our method significantly accelerates the rendering process and, by optionally incorporating sparsely rendered keyframes as guidance, enhances both the physical plausibility and overall visual quality of the output. The resulting pipeline achieves near real-time performance with photorealistic rendering quality, effectively bridging the gap between the efficiency of modern generative models and the precision of traditional physically based rendering. Furthermore, we demonstrate the versatility of our framework by introducing a lightweight, adapter-based module that efficiently repurposes the pretrained forward model for the inverse rendering task of intrinsic decomposition.
\end{abstract}

\section{Introduction}

The pursuit of photorealism in computer graphics has long been driven by
physically based rendering (PBR) techniques, which simulate the complex
interaction of light, materials, and geometry to achieve high visual fidelity
\cite{pharr2023physically,jensen2001realistic}. Methods such as path tracing
are the gold standard for offline applications, such as visual effects in film productions, but their high
computational cost, requiring extensive sampling to obtain ``noise-free''
images, makes them impractical for real-time settings like video games,
interactive virtual production, and pre-visualization, where high frame rates
and rapid creative iteration are essential.

Recent progress in large-scale generative models, especially latent diffusion
models~\cite{rombach2022high}, have enabled neural rendering methods that
synthesize photorealistic images from G-buffer inputs~\cite{zeng2024rgb,
DiffusionRenderer}. Despite their strong visual fidelity, these approaches face
key limitations: the iterative denoising process typically requires 20--50
network evaluations, creating latency unsuitable for interactive use; and the
stochastic sampling inherent to diffusion models reduces physical accuracy and
temporal stability. As a result, flickering and other inconsistencies often
emerge, preventing these methods from meeting studio-quality standards and
limiting their practical deployment.

To bridge the gap between the generative power of diffusion models and the
demands of real-time, physically-consistent rendering, we propose
\emph{RenderFlow}. Our method reformulates neural rendering as a single-step
conditional flow generation process. The core idea is to leverage a flow matching
paradigm~\cite{liu2022flow} to learn a velocity flow field from the diffuse albedo,
rather than Gaussian noise, to the final fully-shaded image, conditioned on the 
G-buffer attributes (e.g., normals, depth, materials). By replacing noise with
albedo as shown in Figure~\ref{fig:teasers}, we preserve low-level signals
while the network synthesizes high-frequency details, maintaining the geometric
integrity of the scene. Learning this direct residual flow enables RenderFlow
to leverage pretrained generative priors for synthesizing complex shading and lighting
effects in a single forward pass.

However, achieving physically accurate results remains challenging for neural
rendering methods, which lack explicit representations of scene geometry and light
transport. To address this and improve both physical accuracy and temporal
stability, we introduce a novel guidance mechanism that leverages sparse, high-quality
reference frames rendered offline via path tracing. These frames provide strong conditioning that anchors the generative process, ensuring the output remains faithful to ground-truth light transport. This optional module effectively reduces visual artifacts and improves consistency, making the method more robust for practical applications.

Furthermore, to demonstrate the versatility of our framework, we introduce a
parameter-efficient adaptation for the inverse rendering task. By augmenting the
frozen \emph{RenderFlow} backbone with adapter modules, our model can be
efficiently repurposed to decompose a rendered image back into its intrinsic
G-buffer components. This addition establishes our approach as a unified
paradigm, capable of handling both forward (synthesis) and inverse
(decomposition) rendering within a single, shared architecture, which can be
potentially applied in further video editing applications.

Our main contributions are:
\begin{itemize}
    \item A single-step, flow-based rendering framework, \emph{RenderFlow}, that
    achieves a faster inference speed by reformulating neural rendering as a
    direct conditional flow generation task, significantly outperforming
    iterative diffusion-based methods in quality and efficiency.
    \item A novel reference-frame guidance module that improves the physical
    accuracy of the rendered output, achieving superior visual quality and
    mitigating artifacts common in prior generative approaches.
    \item A unified, efficient adapter-based inverse renderer built on a frozen forward renderer, demonstrating the versatility and extensibility of \emph{RenderFlow}.
\end{itemize}

\section{Related Work}
\begin{figure*}[t]
\centering
\includegraphics[width=0.9\textwidth]{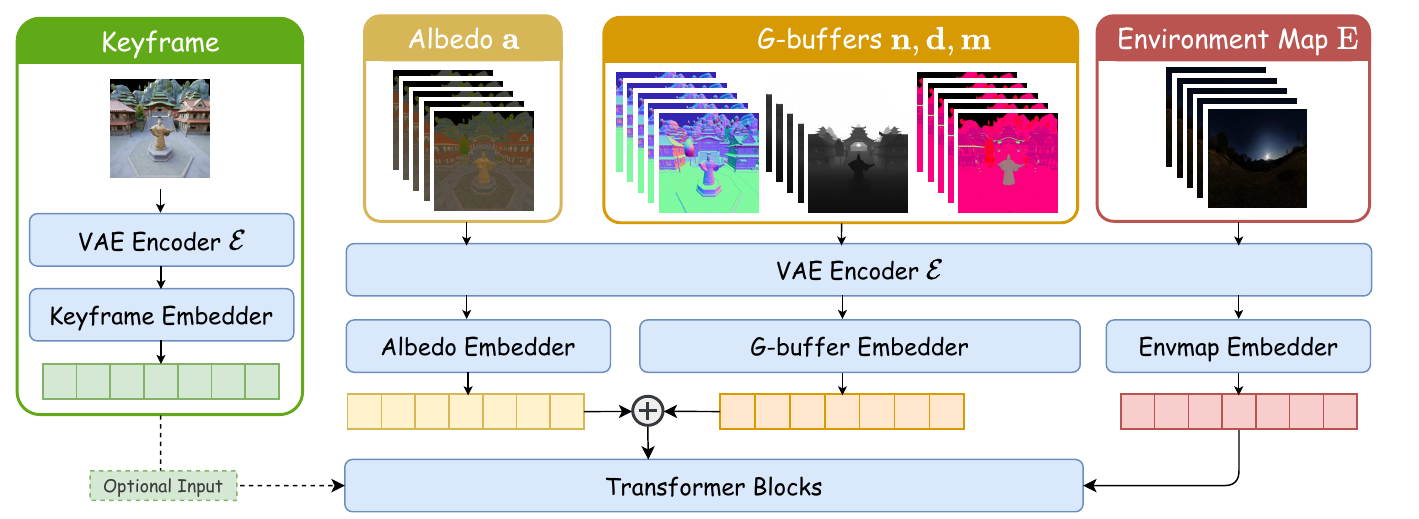}
\vspace{-0.5em}
\caption{\textbf{An overview of our proposed RenderFlow architecture.}
Unlike diffusion methods that start from noise, RenderFlow takes albedo as input and directly predicts fully-shaded outputs.
Built on a pre-trained video DiT, it injects aligned G-buffer tokens into transformer blocks.
Environment maps and optional keyframes are integrated via lightweight adapters (see Fig.~\ref{fig:architecture} for details).
The transformer aggregates all inputs to produce temporally coherent, physically accurate rendering outputs.}
\label{fig:overview}
\vspace{-0.5em}
\end{figure*}

\subsection{Neural Rendering}
Early works employed Convolutional Neural Networks (CNNs) \cite{nalbach2017deep}
and Generative Adversarial Networks (GANs) \cite{thomas2017deep} to approximate
screen-space shading from G-buffers. More recently, Transformer-based
architectures like Lightformer \cite{ren2024lightformer} and Renderformer
\cite{zeng2025renderformer} have been explored to embed lighting and mesh
properties explicitly. The latest generation of methods leverages large-scale
diffusion models for both inverse rendering and forward rendering conditioned on
G-buffer attributes \cite{zeng2024rgb, DiffusionRenderer}, with some works
exploring unified models for joint albedo estimation and relighting
\cite{he2025unirelight}. While these approaches produce visually compelling
results, they are constrained by two critical limitations: the iterative,
multi-step sampling process introduces prohibitive latency for real-time
applications, and the inherent stochasticity of the generative process
compromises the physical fidelity required for production-quality rendering.

\subsection{Efficient Generative Models}
To overcome the latency of iterative sampling, significant research has focused
on accelerating generative models. One major line of work aims to distill
pre-trained, multi-step diffusion models into a single inference step.
Foundational techniques such as adversarial diffusion distillation
\cite{sauer2024adversarial} and distribution matching distillation
\cite{yin2024one} have proven effective at training fast, one-step generators
that maintain high fidelity. These distilled models can be efficiently
fine-tuned for various downstream image-to-image translation tasks such as face
restoration, superresolution, and 3D reconstruction enhancement
\cite{img2img-turbo, zhang2024instantrestore, wu2024one,wu2025difix3d+}.

An alternative and complementary approach is offered by \textbf{Flow Matching}
models, such as Rectified Flow \cite{liu2022flow}. Flow matching learns a velocity vector field that transports
samples from a simple prior distribution such as a Gaussian distribution to the
target data distribution by solving an ordinary differential equation (ODE).
This formulation is highly efficient, as the ODE can be solved with very few
function evaluations. Large-scale flow matching models have demonstrated
state-of-the-art performance in image and video synthesis
\cite{esser2024scaling,labs2025flux1kontextflowmatching,wan2025}. Recent works
have demonstrated the versatility of flow matching in various downstream tasks
including image relighting, shadow generation \cite{chadebec2025lbm}. Our work has similarities
to these advancements, proposing a one-step, flow-based framework
specifically tailored for efficient and high-fidelity neural rendering.

\section{Preliminaries}

\subsection{Classical Rendering Frameworks}
Physically-based rendering (PBR) pipelines approximate this integral with Monte
Carlo path tracing. Although path tracing delivers photorealistic images by
simulating light transport, its heavy computational rules it out for interactive
contexts. Deferred rendering attains real-time performance by decoupling
geometry from shading, where surface attributes (e.g., normals, albedo) are
stored in screen-space G-buffers. A subsequent
lighting pass then uses these buffers for shading, though often at the
cost of the global illumination accuracy provided by path tracing.

\subsection{Diffusion and Bridge Matching Models}

Denoising Diffusion Probabilistic Models (DDPM)~\cite{ho2020denoising} define a
forward Markov process that gradually adds Gaussian noise to data
\(\mathbf{x}_0\), transforming it into a prior distribution. The reverse process
learns to iteratively denoise a sample from the prior back into the data
distribution by training a network \(\epsilon_\theta\) to predict the added
noise at each timestep \(t\).

A limitation of standard diffusion models is their dependence on a fixed
Gaussian prior. The \textbf{bridge matching} framework
\cite{shi2023diffusion,albergo2023stochastic} generalizes this by learning a
stochastic trajectory between two arbitrary distributions \(\pi_0\) and
\(\pi_1\). Given samples \(\mathbf{x}_0 \sim \pi_0\) and \(\mathbf{x}_1 \sim
\pi_1\), the trajectory is defined by a stochastic differential equation (SDE):
\begin{equation}
    \mathrm{d}\mathbf{x}_t = v_\theta(\mathbf{x}_t, t) \mathrm{d}t + \sigma(t) \mathrm{d}\mathbf{W}_t,
    \label{eq:sde}
\end{equation}
where \(v_\theta\) is the learnable velocity (drift) field and
\(\sigma(t)\mathrm{d}\mathbf{W}_t\) adds stochastic perturbation via a
time-dependent noise schedule, \(\sigma(t)\). A common choice, the \textit{Brownian bridge},
uses \(\sigma(t) = \sigma \sqrt{t(1 - t)}\) for smooth interpolation. The
velocity field is trained with:
\begin{equation}
    \mathcal{L}_{\text{Bridge}} = \mathbb{E}_{\mathbf{x}_0, \mathbf{x}_1, t}
    \left[ \left\| v_\theta(\mathbf{x}_t, t) - \frac{\mathbf{x}_1 - \mathbf{x}_t}{1 - t} \right\|^2 \right],
\end{equation}
In the deterministic case where \(\sigma(t) = 0\), this reduces to
\textbf{flow matching}, which learns a straight-line ODE path:
\begin{equation}
    \mathcal{L}_{\text{Flow}} = \mathbb{E}_{\mathbf{x}_0, \mathbf{x}_1, t}
    \left[ ||v_\theta(\mathbf{x}_t, t) - (\mathbf{x}_1 - \mathbf{x}_0)||^2 \right],
\end{equation}
where \(\mathbf{x}_t = (1-t)\mathbf{x}_0 + t\mathbf{x}_1\).

\section{Method}
We formulate neural rendering as a conditional flow-based generative modeling problem,
learning a transformation from structured G-buffer attributes to the
distribution of photorealistic rendered sequences. Leveraging the powerful prior of a large-scale pretrained video diffusion model, we reinterpret its
dynamics as a conditional latent flow. We introduce
a novel architecture and a training scheme based on the principles of bridge
matching \cite{liu2022flow}, enabling high-fidelity rendering in a
single inference step.

\begin{figure*}[t]
    \centering
    \includegraphics[width=0.9\textwidth]{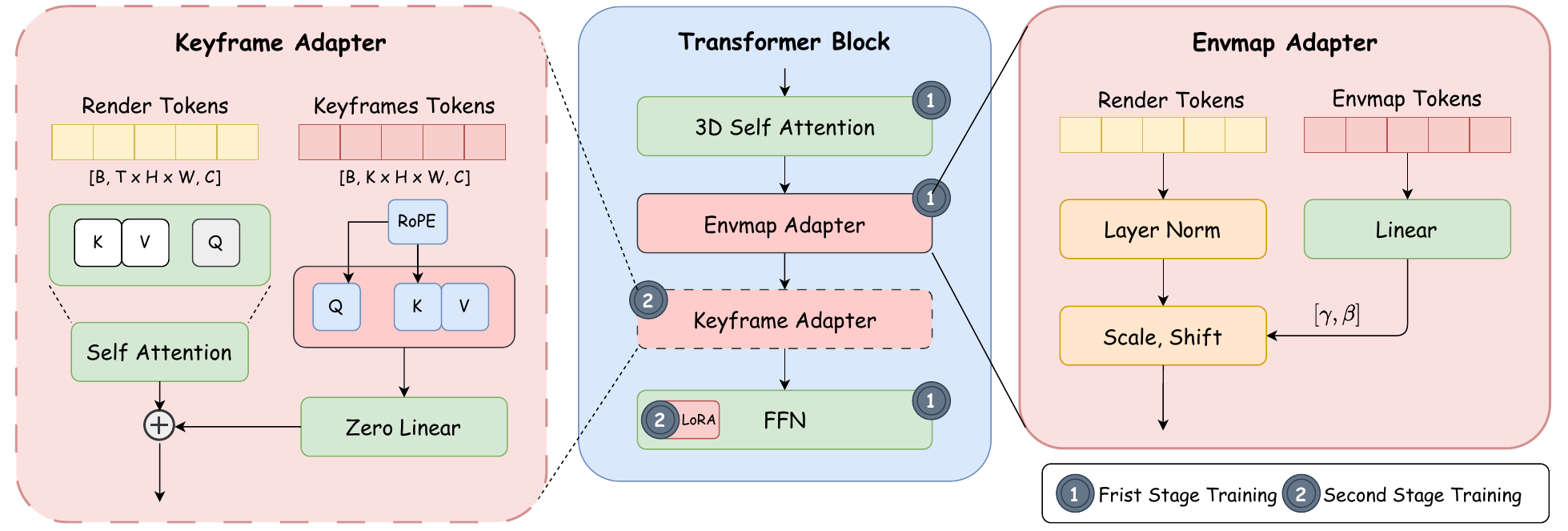}
    \vspace{-0.5em}
    \caption{\textbf{Design of the customized transformer block in RenderFlow.}
        To incorporate global lighting context, the environment map is injected via an adaptive normalization layer before the transformer layers.
        To compensate for the limited information in per-frame G-buffers, the keyframe adapter introduces cross-attention to extract complementary temporal features from an \emph{optional} keyframe.
        The adapter includes a LoRA module for parameter-efficient adaptation and employs Rotary Position Embeddings (RoPE) to encode temporal distances between keyframes and current frames.}
    \label{fig:architecture}
    \vspace{-0.5em}
\end{figure*}

\subsection{Input Representation}
Following the deferred rendering paradigm, our model processes a set of
G-buffers. The primary input is the \textbf{Albedo} buffer \(\mathbf{a} \in
\mathbb{R}^{H \times W \times 3}\), which provides the base color. Geometric
information is supplied by the \textbf{Normal} buffer \(\mathbf{n} \in
\mathbb{R}^{H \times W \times 3}\) (in camera space) and a single-channel
\textbf{Depth} buffer \(\mathbf{d} \in \mathbb{R}^{H \times W}\). Material
properties, adhering to the widely adopted Disney BRDF principle
\cite{burley2012physically, karisRealShadingUnreal2013}, are described by three
single-channel buffers: roughness, metallic, and specular. These are
concatenated along the channel axis to form a single three-channel
\textbf{Material} buffer \(\mathbf{m}\).

For lighting conditions, we follow prior work
\cite{DiffusionRenderer} and represent illumination with a
high-dynamic-range (HDR) environment map \(\mathbf{E} \in \mathbb{R}^{H' \times
    W' \times 3}\). By contrast, we simplify the input representation by rotating
the environment map into the camera's view space and applying Reinhard tone
mapping operator \cite{reinhard2023photographic} to produce a single LDR image,
\(\mathbf{E}_{\text{ldr}}\). We hypothesize that, at high resolution, a tone
mapped environment map is sufficient for capturing the lighting information,
making extra intensity channels in other dynamic ranges unnecessary. Rotating
the map to the camera view during training allows the network to model the
directional lighting implicitly, without explicit directional encodings.

\subsection{Model Architecture}
\textbf{Backbone.} Our architecture is built upon a pre-trained video diffusion
model, Wan2.1 \cite{wan2025}, which features a 3D causal VAE and a Diffusion
Transformer (DiT) backbone. As illustrated in Fig. \ref{fig:overview}, all G-buffer attributes and the tone mapped
environment map are first processed by the VAE encoder to obtain latent
representations: \(\mathcal{E}(\mathbf{a}), \mathcal{E}(\mathbf{n}),
\mathcal{E}(\mathbf{d}), \mathcal{E}(\mathbf{m}),
\mathcal{E}(\mathbf{E}_\text{ldr})\). To repurpose the model for our task, we
replace the standard noisy latent input with a clean albedo latent
\(\mathcal{E}(\mathbf{a})\). An input embedder then patchifies the albedo
latent into render tokens. Following VACE \cite{jiang2025vace}, the remaining
spatially aligned G-buffer latents are processed by a decoupled attribute
embedder. Given their spatial alignment, the resulting attribute tokens are
added element-wise to the render tokens before being fed into the transformer
blocks.

\textbf{Envmap Adapter.} Unlike the G-buffers, the environment map
\(\mathbf{E}_{\text{ldr}}\) lacks spatial alignment with the render tokens.
Instead, it functions as a global condition, governing the overall color
palette and lighting characteristics of the scene. To inject this global
information, we modulate the render tokens within each transformer block with an adaptive normalization layer
 \cite{xu2019understanding}. Specifically,
the latent representation of the environment map is processed into envmap
tokens. These tokens are then passed through a linear projection to predict the
scale factor (\(\gamma\)) and shift (\(\beta\)) parameter. The projected render features are defined as the linear projection $f_{i+1}=(\gamma + 1) * f_{i} + \beta$, where
$f_i$ represents the render features at layer $i$.

\textbf{Keyframe Guidance.} To enable optional sparse-keyframe guidance for
improved physical accuracy and visual quality, we introduce a Keyframe Adapter
module (Fig.~\ref{fig:architecture}). This module adds a cross-attention
branch in parallel to the existing self-attention layer, whose output is added to the original feature as a residual correction term. To encode the temporal distance between the current
frame and each keyframe, we apply Rotary Position Embeddings (RoPE)
\cite{su2024roformer} to the key and query vectors. Since this adapter is an
optional component, we adopt a two-stage training strategy: Stage 1 trains the
base model alone to learn the core rendering task, and Stage 2 freezes it while
training only the Keyframe Adapter. This ensures that baseline performance
remains unchanged when no keyframes are provided.

\subsection{Latent-Space Bridge-Matching Training}
We train our model under the stochastic bridge matching framework
\cite{albergo2023stochastic}, aiming to learn a conditional trajectory that transforms
the input albedo distribution to the distribution of fully rendered image. For computational
efficiency, we formulate this process in the latent space. Let \(\mathbf{z}_0 =
\mathcal{E}(\mathbf{a})\) be the latent of the input albedo and \(\mathbf{z}_1 =
\mathcal{E}(\mathbf{I}_{\text{gt}})\) be the latent of the ground truth image.
During training, we sample a timestep \(t \sim \pi(t)\) where \(t \in [0, 1]\)
and interpolate between the start and end points of the path using the Brownian
bridge formulation:
\begin{equation}
    \mathbf{z}_t = (1 - t) \mathbf{z}_0 + t \mathbf{z}_1 + \sigma \sqrt{t(1 - t)} \epsilon
    \label{eq:bridge_interpolation}
\end{equation}
where \(\epsilon \sim \mathcal{N}(0, \mathbf{I})\). The network
takes the interpolated latent \(\mathbf{z}_t\), the timestep \(t\), and
conditioning information as input to predict the conditional \(v_\theta\). The
model is trained to minimize the bridge matching loss in the latent space:
\begin{equation}
    \mathcal{L}_{\text{latent}} = \mathbb{E}_{\mathbf{z}_0, \mathbf{z}_1, t}
    \left[ \left\| v_\theta(\mathbf{z}_t, t) - \frac{\mathbf{z}_1 - \mathbf{z}_t}{1 - t} \right\|^2 \right]
    \label{eq:latent_loss}
\end{equation}
Crucially, with a trained velocity field, \(v_\theta\), the final latent
\(\hat{\mathbf{z}}_1\) can be derived directly from the interpolated latent
\(\mathbf{z}_t\):
\begin{equation}
    \hat{\mathbf{z}}_1 = \mathbf{z}_t + v_\theta(\mathbf{z}_t, t) (1 - t).
\end{equation}
Inspired by image translation literature \cite{img2img-turbo,chadebec2025lbm},
we decode the predicted latent \(\hat{\mathbf{z}}_1\) to the final rendered
image \(\mathbf{I}_{\text{pred}} = \mathcal{D}(\hat{\mathbf{z}}_1)\) and apply
additional pixel-space losses to accelerate convergence and enhance reconstruction
fidelity. Specifically, we optimize a composite pixel-wise loss:
\begin{equation}
    \mathcal{L}_{\text{pixel}} = \mathcal{L}_{\text{LPIPS}} + \mathcal{L}_{\text{grad}},
    \label{eq:pixel_loss}
\end{equation}
which combines the LPIPS loss \cite{zhang2018perceptual} for perceptual
similarity and a gradient loss \cite{ma2020structure} to reconstruct
high-frequency details like contact shadows. The final training objective is a
weighted sum of the latent and pixel losses:
\begin{equation}
    \mathcal{L}_{\text{total}} = \mathcal{L}_{\text{latent}} + \lambda \mathcal{L}_{\text{pixel}}.
    \label{eq:total_loss}
\end{equation}

\subsection{Long video inference}
\label{subsec:long_video_inf}
Using pixel-space losses requires decoding latent representations at every
training step, which is computationally expensive for long video sequences.
We therefore train the model on short, fixed-length clips (typically 5 frames)
and employ a progressive inference strategy for long videos. Specifically,
during training, we randomly condition the model on the first frame of the
input sequence and predict subsequent frames by feeding a masked reference clip,
following VACE \cite{jiang2025vace}. At inference time, we generate videos in
overlapping chunks, where the last frame of the previous chunk serves as the
conditioning frame for the next, promoting smooth transitions and temporal
coherence over the full sequence.

\subsection{Inverse Rendering}
\begin{figure}[t]
       \centering
       \includegraphics[width=\columnwidth]{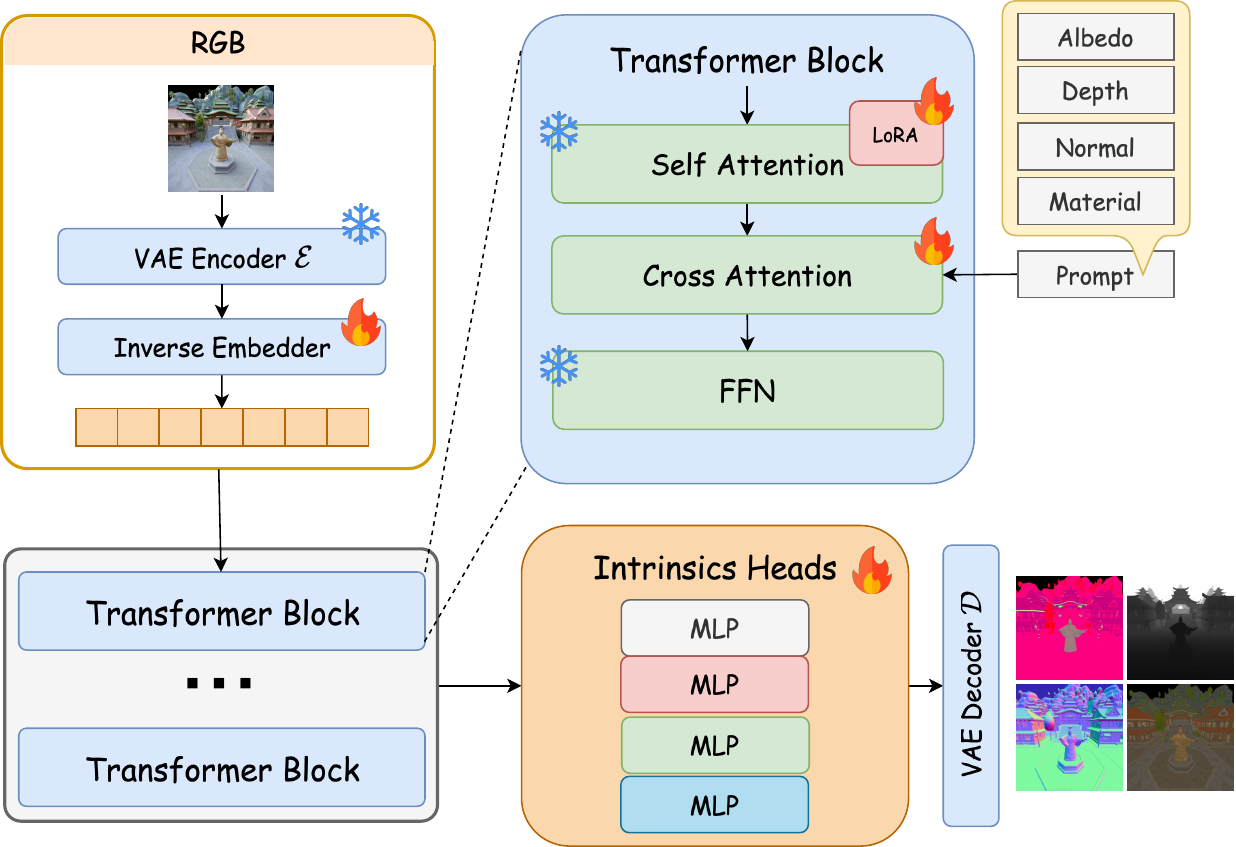}
       \vspace{-1em}
       \caption{\textbf{Inverse adapter architecture.} The inverse adapter uses
       the frozen forward rendering backbone and introduces specialized modules
       to adapt it for G-buffer decomposition.}
       \label{tab:inverse_adapter}
       \vspace{-1em}
\end{figure}
\begin{figure*}[htbp]
    \centering
    \includegraphics[width=\textwidth]{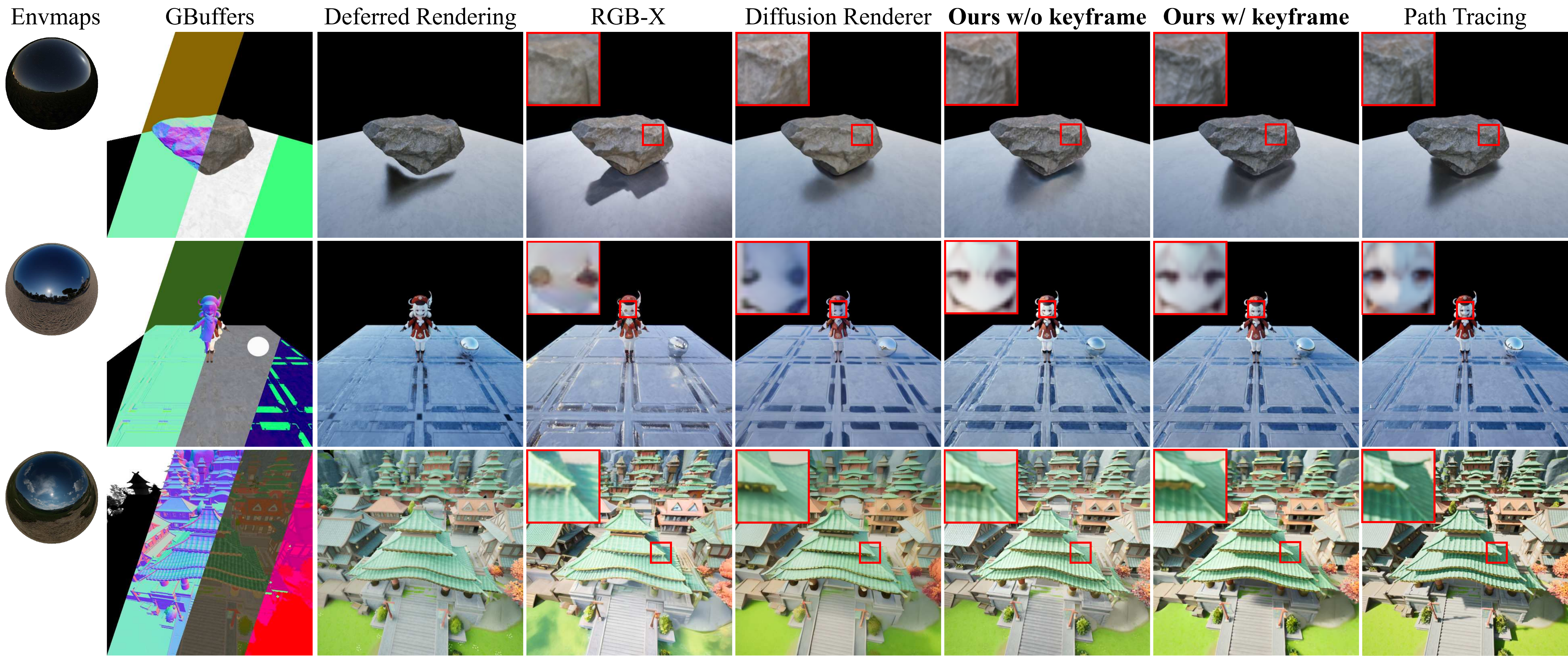}
    \label{tab:qualitative_visualizations}
    \vspace{-1em}
    \caption{\textbf{Qualitative comparison of rendered images with traditional methods
        including deferred rendering, and
        diffusion-based approaches (RGB-X and DiffusionRenderer)}. Our method
        better preserves texture details and produces high-quality shadows and
        lighting effects. }    
\end{figure*}
Our framework can be effectively repurposed for inverse rendering tasks without
training a separate model from scratch. Concretely, we freeze the pretrained
forward backbone and introduce a lightweight inverse adapter. An input RGB image
is first encoded by the frozen VAE encoder $\mathcal{E}$ into a latent
$z_{\text{rgb}}$, which is then mapped to tokens by a trainable inverse embedder. To repurpose the frozen transformer for intrinsic decomposition, we augment its self-attention projections with low-rank
adaptation (LoRA) modules and insert a prompt-conditioned cross-attention branch where a text prompt
selects the target intrinsic to be predicted. After the final transformer block, a set of
lightweight, per-intrinsic heads (MLPs) project the tokens into the
corresponding latent space, which is then decoded by the frozen VAE decoder
$\mathcal{D}$ to produce the requested G-buffer. This design preserves the
representational rendering features of the forward model, enables parameter-efficient
adaptation, and unifies multiple inverse tasks within a single model switchable
by prompt, as illustrated in Fig. \ref{tab:inverse_adapter}.

Training optimizes only the adapter parameters (inverse embedder, LoRA,
cross-attention, and per-intrinsic heads) using modality specific reconstruction objectives
: L1+LPIPS for albedo, cosine similarity for normals, scale-and
shift-invariant loss for depth \cite{ranftl2020towards}, and L1 for material
components. In practice, this yields accurate and stable decomposition while
retaining the forward renderer unchanged and avoiding full model retraining.

\subsection{Dataset Synthesis}
Due to the lack of a large-scale rendering dataset that contains the buffers and
environment maps required in our task, we constructed a large-scale dataset
using Unreal Engine 5 to train our model. Leveraging the \textit{Movie Render
Queue}, we rendered high-quality, path-traced reference images alongside their
corresponding G-buffers. The dataset is composed of two distinct categories:
large-scale, artist-crafted environments, and procedurally generated synthetic scenes. To overcome the
computational expense and limited availability of complex, pre-existing scenes,
we developed a procedural pipeline to compose novel scenes from a curated
collection of assets, significantly increasing data diversity. This collection
includes approximately 4,000 unique meshes and 30 high-dynamic-range imaging
(HDRI) environment maps. To further enrich the dataset, we randomly configured
the material attributes of the meshes. In total, our dataset contains over
30,000 frames from artist-crafted environments and 100,000 frames from synthetic
scenes. All images were rendered at a 512x512 resolution with 256 samples per
pixel (SPP) and subsequently denoised using Intel's Open Image Denoise
\cite{OpenImageDenoise}.

\section{Experiments}

We refer to implementation details in the Supplementary.

\begin{figure}[ht]
       \centering
       \includegraphics[width=\columnwidth]{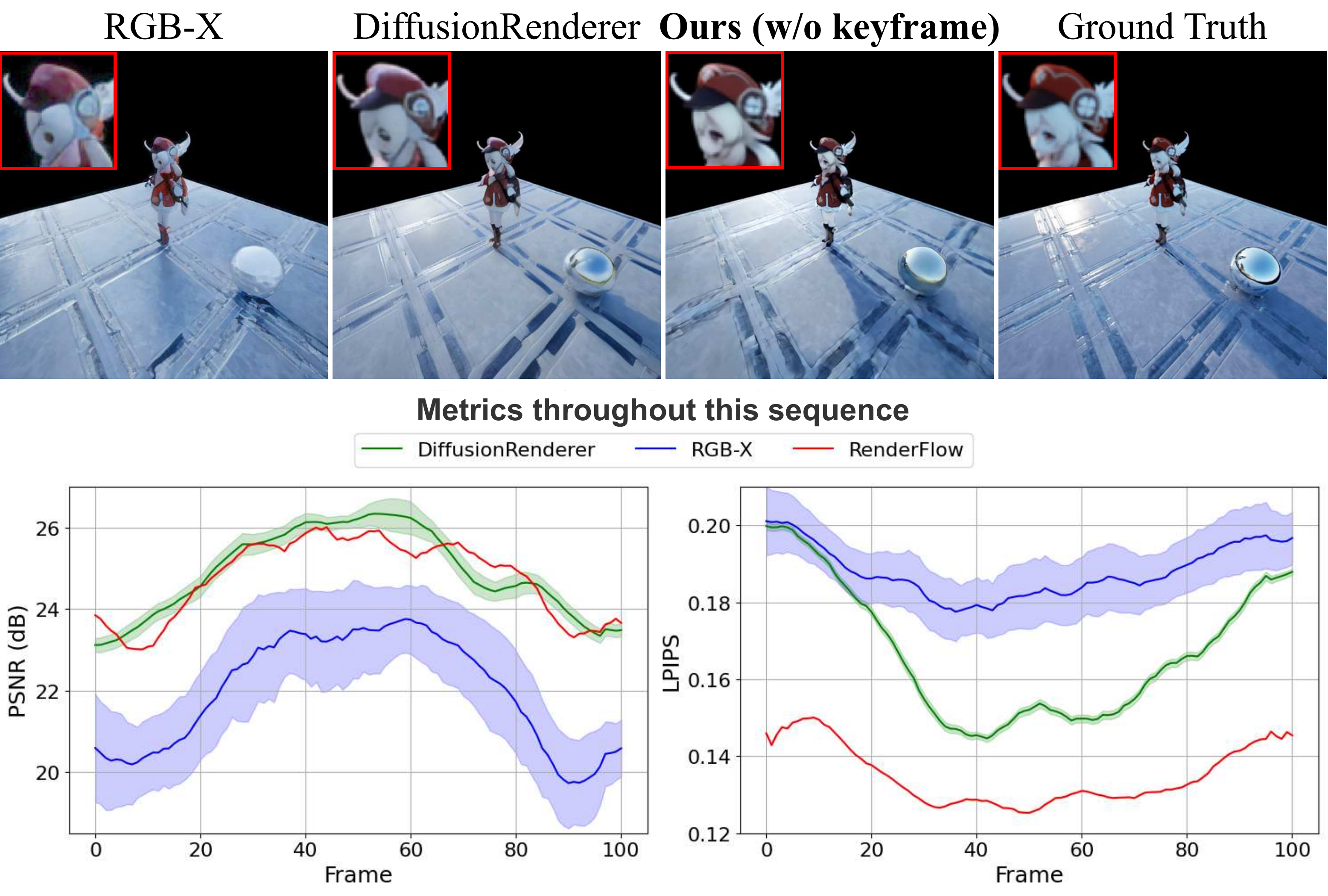}
       \caption{\textbf{Metrics are evaluated over a 100-frame sequence and averaged across 10 runs.} Shaded regions indicate variance across runs. Our method not only achieves the best overall performance but also zero variance, reflecting its deterministic, consistently high-quality behavior in contrast to the stochastic baselines.}

       \label{fig:metrics_variance}
       \vspace{-1em}
\end{figure}
\subsection{Evaluation}
\begin{figure*}[htbp]
    \centering
    \includegraphics[width=\textwidth]{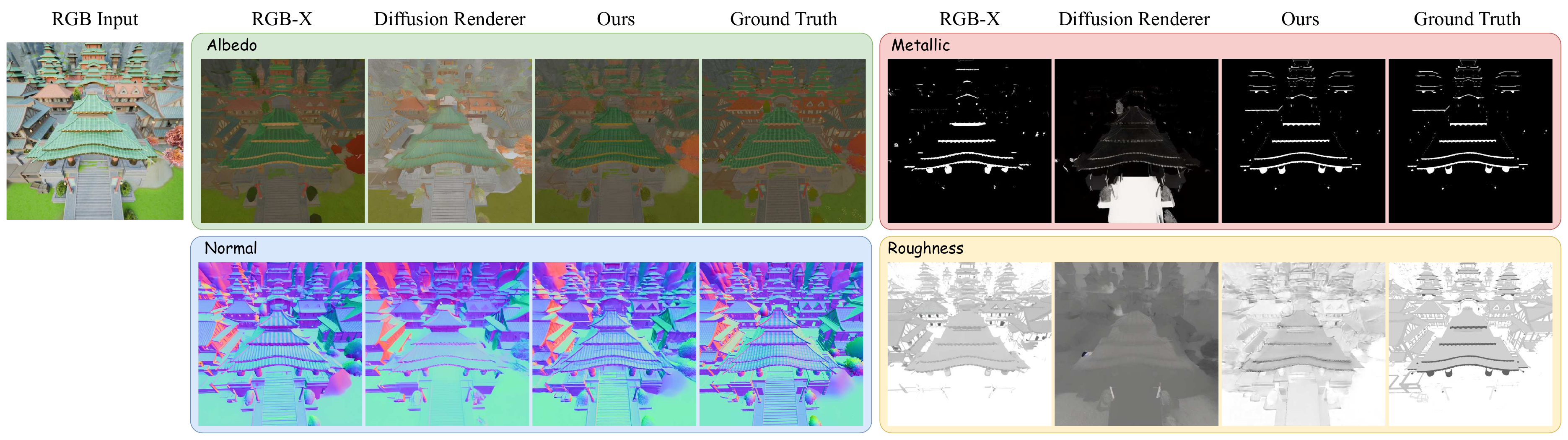}
    \caption{\textbf{Qualitative comparison of intrinsic decomposition.} We compare our adapter-based method against RGB-X and DiffusionRenderer. This figure demonstrates the decomposition results for key G-buffer components: \textcolor{green}{Albedo}, \textcolor{blue}{Normal}, \textcolor{red}{Metallic}, and \textcolor{yellow}{Roughness}.}
    \vspace{-1em}
    \label{tab:inverse_visualizations}
\end{figure*}

\textbf{Baselines.} We compare against two state-of-the-art
diffusion-based neural rendering frameworks: RGB-X \cite{zeng2024rgb} and
DiffusionRenderer \cite{DiffusionRenderer}. To ensure a fair comparison and
mitigate potential domain gaps, we fine-tuned both baselines on our own dataset,
following the training setups in their original papers.
RGB-X was fine-tuned for 100,000 steps, while DiffusionRenderer, which operates
on video frames, was fine-tuned on 4-frame sequences for 25,000 steps. At
inference, we use 50 sampling steps for RGB-X and 30 steps for DiffusionRenderer.

\textbf{Test Set and Metrics.} We curated a challenging test set comprising both
synthetic and artist-crafted scenes featuring complex lighting and shadows. The
set includes 1,000 paired frames for synthetic scenes and 1,000 frames for
artist-crafted scenes. We evaluate all methods using standard image
reconstruction metrics: Peak Signal-to-Noise Ratio (PSNR), Structural Similarity
Index (SSIM), and the Learned Perceptual Image Patch Similarity (LPIPS)
\cite{zhang2018perceptual}.

\begin{table}[t]
    \centering
    \footnotesize
    \setlength{\tabcolsep}{0.8pt} %
    \begin{tabular}{l cc ccc c}
        \toprule
        \textbf{Method} & \textbf{Paradigm} & \textbf{Params} & \textbf{PSNR} $\uparrow$ & \textbf{SSIM} $\uparrow$ & \textbf{LPIPS} $\downarrow$ & \textbf{Time (s)} $\downarrow$ \\
        \midrule
        \textcolor{gray}{Path Tracing} & \textcolor{gray}{Traditional} & \textcolor{gray}{-} & \textcolor{gray}{-} & \textcolor{gray}{-} & \textcolor{gray}{-} & \textcolor{gray}{$>$10} \\
        \textcolor{gray}{IBL} & \textcolor{gray}{Traditional} & \textcolor{gray}{-} & \textcolor{gray}{16.903} & \textcolor{gray}{0.882} & \textcolor{gray}{0.105} & \textcolor{gray}{-} \\
        \textcolor{gray}{Deferred} & \textcolor{gray}{Traditional} & \textcolor{gray}{-} & \textcolor{gray}{24.649} & \textcolor{gray}{0.927} & \textcolor{gray}{0.097} & \textcolor{gray}{-} \\
        \midrule
        RGB-X \cite{zeng2024rgb} & Diffusion & 950M & 20.984 & 0.793 & 0.165 & $\sim$2.19 \\
        DRender \cite{DiffusionRenderer} & Diffusion & 1.7B & 23.758 & 0.863 & 0.128 & $\sim$1.40 \\
        \midrule
        Ours (w/o key) & Flow & 1.4B & \underline{24.214} & \underline{0.874} & \underline{0.113} & \textbf{$\sim$0.19} \\
        Ours (w/ key) & Flow & 1.7B & \textbf{26.663} & \textbf{0.883} & \textbf{0.101} & \underline{$\sim$0.24} \\
        \bottomrule
    \end{tabular}
    \caption{\textbf{Quantitative comparison (512x512).}
        Our method achieves SOTA quality, while being $\sim$10x faster
        than diffusion baselines. Keyframe guidance further improves fidelity
        with negligible speed impact.}
    \label{tab:quantitative_results}
    \vspace{-1em}
\end{table}
\textbf{Quantitative Results.} As shown in Tab. \ref{tab:quantitative_results},
our single-step method achieves superior performance to existing diffusion-based methods. Without keyframe
guidance, it outperforms both RGB-X and DiffusionRenderer across all metrics. 
When keyframe guidance is enabled, the full model attains the best overall
performance, with substantial improvements on all metrics, confirming
the effectiveness of the proposed guidance mechanism. Qualitative visualizations
in Fig.\ref{tab:quantitative_results} further support these findings. We also report results from Unreal
Engine's native deferred rendering pipeline. While our learned model is not intended as a direct replacement for
highly-optimized industry pipelines that operate on explicit
high-polygon geometry, the experimental results show that our proposed learning based approach can
faithfully reproduce challenging effects such as reflections that are expensive in
real-time rendering pipelines. A critical advantage of our approach is its deterministic
nature: as illustrated in Fig.\ref{fig:metrics_variance}, our method exhibits
zero variance across multiple inference runs, in contrast to
the stochastic diffusion baselines, which show significant variance in their
outputs. This determinism yields consistent and reproducible
output, making our method more reliable for production environments
and other applications requiring stable, predictable outputs.

\textbf{Inference Speed.} We also highlight the significant efficiency gains of
our method over both iterative baselines and traditional path tracing in Tab.
\ref{tab:quantitative_results}. While path tracing times can range from seconds
to hours depending on scene complexity, our image-space approach is decoupled
from scene geometry, ensuring a consistent inference speed. Compared to
diffusion-based methods, our single-step model renders a 512x512 frame in
approximately 0.19 seconds on a single NVIDIA RTX 4090 GPU, making it over 7x
faster than DiffusionRenderer and more than 10x faster than RGB-X. Insight: the primary
computational overhead stems from the VAE, with G-buffer encoding and final
image decoding taking approximately 0.12s and 0.04s respectively, together
accounting for nearly 90\% of the total inference time.

\textbf{Inverse Rendering.} We also evaluate our inverse adapter's ability to
decompose images into their underlying G-buffers. Note that we are not intended
to defeat specialized inverse rendering methods, but to provide a unified
framework for forward and inverse rendering. As shown in Tab.
\ref{tab:inverse_rendering_results} and Fig. \ref{tab:inverse_visualizations}, our method demonstrates competitive
performance in intrinsics decomposition compared to RGB-X and
DiffusionRenderer, highlighting its versatility.

\begin{table}[t]
       \centering
       \setlength{\tabcolsep}{1pt}
       \begin{tabular}{l cc c c}
              \toprule
                                               & \multicolumn{2}{c}{Albedo} & \multicolumn{1}{c}{Normal}  & \multicolumn{1}{c}{Material}                                        \\
              \cmidrule(lr){2-3}\cmidrule(lr){4-4}\cmidrule(lr){5-5}
              \textbf{Method}                  & \textbf{PSNR} $\uparrow$   & \textbf{LPIPS} $\downarrow$ & \textbf{Angular} ($^\circ$) $\downarrow$ & \textbf{RMSE} $\downarrow$ \\
              \midrule
              RGB-X \cite{zeng2024rgb}         & 26.24                      & \textbf{0.116}              & 46.5                                   & \textbf{0.080}             \\
              DRender \cite{DiffusionRenderer} & 20.78                      & 0.182                       & 47.6                                  & 0.194                      \\
              Ours                             & \textbf{26.93}             & 0.153                       & \textbf{16.2}                          & 0.084                      \\
              \bottomrule
       \end{tabular}
       \caption{\textbf{Quantitative comparison of inverse rendering.} We
              compare our method with RGB-X and DiffusionRenderer on intrinsics
              decomposition. The results demonstrate that our
              parameter-efficient approach achieves competitive performance
              across key G-buffer attributes.}
        \vspace{-0.5cm}
       \label{tab:inverse_rendering_results}
\end{table}

\subsection{Ablation Study}

We conduct ablation studies to evaluate the effectiveness of the training
strategies of bridge matching, the pixel losses, and the keyframe guidance. All
the experiments are conducted on lower resolution (256x256) to reduce the
computational cost and trained on \(4 \times\) RTX 4090s.

\textbf{Bridge Matching.} We compare different training strategies, including
flow matching (ODE) and bridge matching (SDE). Following previous fast diffusion
works \cite{chadebec2025flash,sauer2024fast}, we compare the model trained with
a discrete timesteps sampling scheme for fixed four timesteps \([1.0, 0.75,
              0.5,0.25]\) or in a uniform timesteps. For the SDE-based approach, we follow LBM
\cite{liu2022flow} and set the bridge noise \(\sigma\) to 0.005. As shown in
Tab. \ref{tab:training_strategy}, we empirically find that single-step
inference yields superior results for models trained with multi-step schedules,
which we attribute to the avoidance of error propagation. Insights: the noise perturbation
in the SDE formulation (Eq. \ref{eq:bridge_interpolation}) encourages the model
to generate more diverse effects, thereby increasing its robustness. Ultimately,
training with a 4-step SDE schedule and performing inference in a single step
provides the best balance of quality and robustness.

\textbf{Pixel Losses.} We evaluate the impact of the pixel-space losses defined
in Eq. \ref{eq:pixel_loss}. As detailed in Tab. \ref{tab:qualitative_image_loss}, incorporating an LPIPS loss alongside the
latent-space loss significantly improves perceptual quality. The addition of a
gradient loss further enhances high-frequency details, such as contact shadows,
which are crucial for realistic rendering, leading to the best overall PSNR and
SSIM scores.

\begin{figure}[t]
       \centering
       \includegraphics[width=\columnwidth]{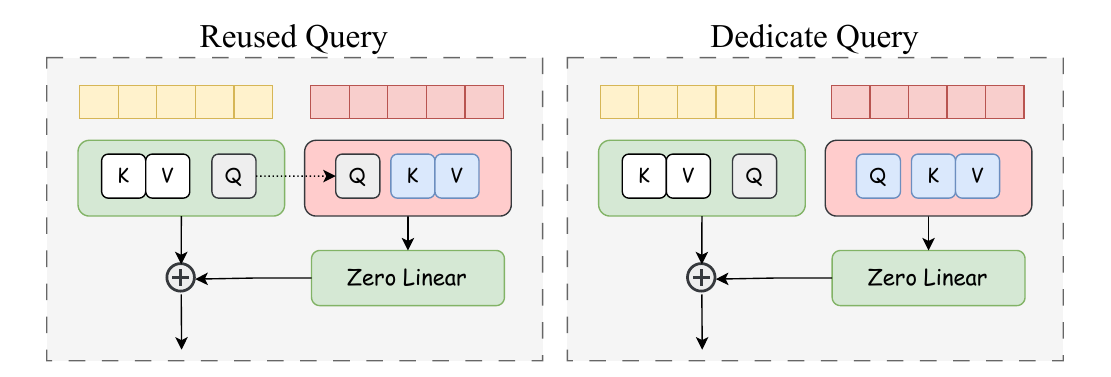}
       \caption{\textbf{Two different keyframe adapter designs.} The left one reuses the
              query from the self-attention layer for efficiency, while the
              right one uses a dedicated query.}
       \label{tab:keyframe_adapter}
       \vspace{-1em}
\end{figure}
\textbf{Keyframe Guidance.} We compare several different keyframe adapter designs. We begin with exploring two attention mechanisms with similar ideas to \cite{jiang2023res} as shown in Fig.
\ref{tab:keyframe_adapter}. The first option is to reuse the query from the
self-attention layer, which is the most efficient method of tuning the model for
keyframes. Another choice is to use a dedicated query (a full attention layer)
for keyframes, which is more flexible but requires more parameters. Moreover, we
empirically find that adding a LoRA \cite{hu2022lora} layer in the feedforward
layer (ffn) in the transformer block can further improve the performance, as it allows
the model to adapt to the fused keyframe features. Tab.
\ref{tab:qualitative_keyframes} shows that the dedicated query with LoRA achieves
the best performance, outperforming the other two designs. Moreover, Tab.
\ref{tab:qualitative_keyframes}, \textit{VACE progressive}, shows
that only applying the progressive inference strategy is not sufficient, as the
keyframe information is propagated across chunks instead of being directly fed
in. Instead, our keyframe adapter provides a more flexible and effective way to
incorporate keyframe information, leading to significantly improved rendering
quality. The results demonstrate that our method achieves the best performance
when combined with the progressive inference strategy. All the experiments are 
conducted with a keyframe gap of 16 frames. Further analysis on the influence
of keyframe gaps is elaborated in the supplementary materials.

\begin{table}[t]
       \centering
       \setlength{\tabcolsep}{3pt}
       \begin{tabular}{l|ccc}
              \toprule
              \textbf{Method}           & PSNR $\uparrow$ & SSIM $\uparrow$ & LPIPS $\downarrow$ \\
              \midrule
              Uniform SDE (4 steps)     & 22.192          & 0.858           & 0.120              \\
              4 timesteps ODE (4 steps) & 23.089          & 0.865           & 0.110              \\
              4 timesteps ODE (1 step)  & 23.304          & 0.867           & 0.108              \\
              4 timesteps SDE (4 steps) & 23.384          & 0.865           & 0.111              \\
              4 timesteps SDE (1 step)  & \textbf{23.590} & \textbf{0.868}  & \textbf{0.107}     \\
              \bottomrule
       \end{tabular}
       \caption{\textbf{Quantitative evaluation of different training strategies.} The
              method name describes the training and inference setup (e.g., ``4 timesteps SDE (1 step)''
              denotes training with a 4-step SDE schedule but performing inference in a single step).}
       \label{tab:training_strategy}
\end{table}

\begin{table}[t]
       \centering
       \begin{tabular}{l|ccc}
              \toprule
              \textbf{Method}                                            & PSNR $\uparrow$ & SSIM $\uparrow$ & LPIPS $\downarrow$ \\
              \midrule
              \(L_{\text{latent}}\) only                                 & 21.588          & 0.840           & 0.148              \\
              \(L_{\text{latent}} + L_{\text{lpips}}\)                   & 23.538          & 0.867           & \textbf{0.105}     \\
              \(L_{\text{latent}} + L_{\text{lpips}} + L_{\text{grad}}\) & \textbf{23.590} & \textbf{0.868}  & 0.107              \\
              \bottomrule
       \end{tabular}
       \caption{\textbf{Quantitative evaluation of the pixel losses.} Experiments demonstrate
              that the LPIPS loss significantly improves perceptual quality, and
              the gradient loss further enhances high-frequency details.}
       \label{tab:qualitative_image_loss}
\end{table}

\begin{table}[t]
       \centering
       \setlength{\tabcolsep}{3pt} %
       \begin{tabular}{l|ccc}
              \toprule
              \textbf{Method}                  & PSNR $\uparrow$ & SSIM $\uparrow$ & LPIPS $\downarrow$ \\
              \midrule
              w/o Keyframes                    & 23.590          & 0.868           & 0.107              \\
              \midrule
              VACE progressive                 & 25.042          & 0.877           & 0.102              \\
              Reusd Query w/o ffn lora         & 24.520          & 0.862           & 0.098              \\
              Dedicated Query w/o ffn lora     & 25.217          & 0.878           & 0.097              \\
              Dedicated Query w/ ffn lora      & 25.720          & 0.881           & 0.094              \\
              \textbf{Ours + VACE progressive} & \textbf{26.220} & \textbf{0.885}  & \textbf{0.090}     \\
              \bottomrule
       \end{tabular}
       \caption{\textbf{Quantitative evaluation of the Keyframe Adapter designs.} A dedicated attention query with LoRA in the feedforward layer outperforms the reused one. When combined with the progressive strategies (Sec.~\ref{subsec:long_video_inf}), our full design achieves the highest overall metrics (last row).}
       \vspace{-0.5cm}
       \label{tab:qualitative_keyframes}
\end{table}

\section{Conclusion}

In this paper, we introduce \emph{RenderFlow}, a novel framework for high-fidelity
neural rendering that operates in a single diffusion step. By reformulating the rendering
as a conditional flow matching problem, our model learns a direct mapping from
G-buffer inputs to the distribution of final images, bypassing the
iterative and stochastic sampling procedures of standard diffusion models. When this approach
is augmented with a novel sparse keyframe guidance mechanism, RenderFlow attains higher
visual fidelity than diffusion-based neural rendering baselines while
substantially reducing inference time, making it a promising solution for real-time and
interactive applications.

{
    \small
    \bibliographystyle{ieeenat_fullname}
    \bibliography{main}
}

\clearpage
\appendix

\maketitlesupplementary

\begin{figure*}[htbp]
    \centering
    \includegraphics[width=\linewidth]{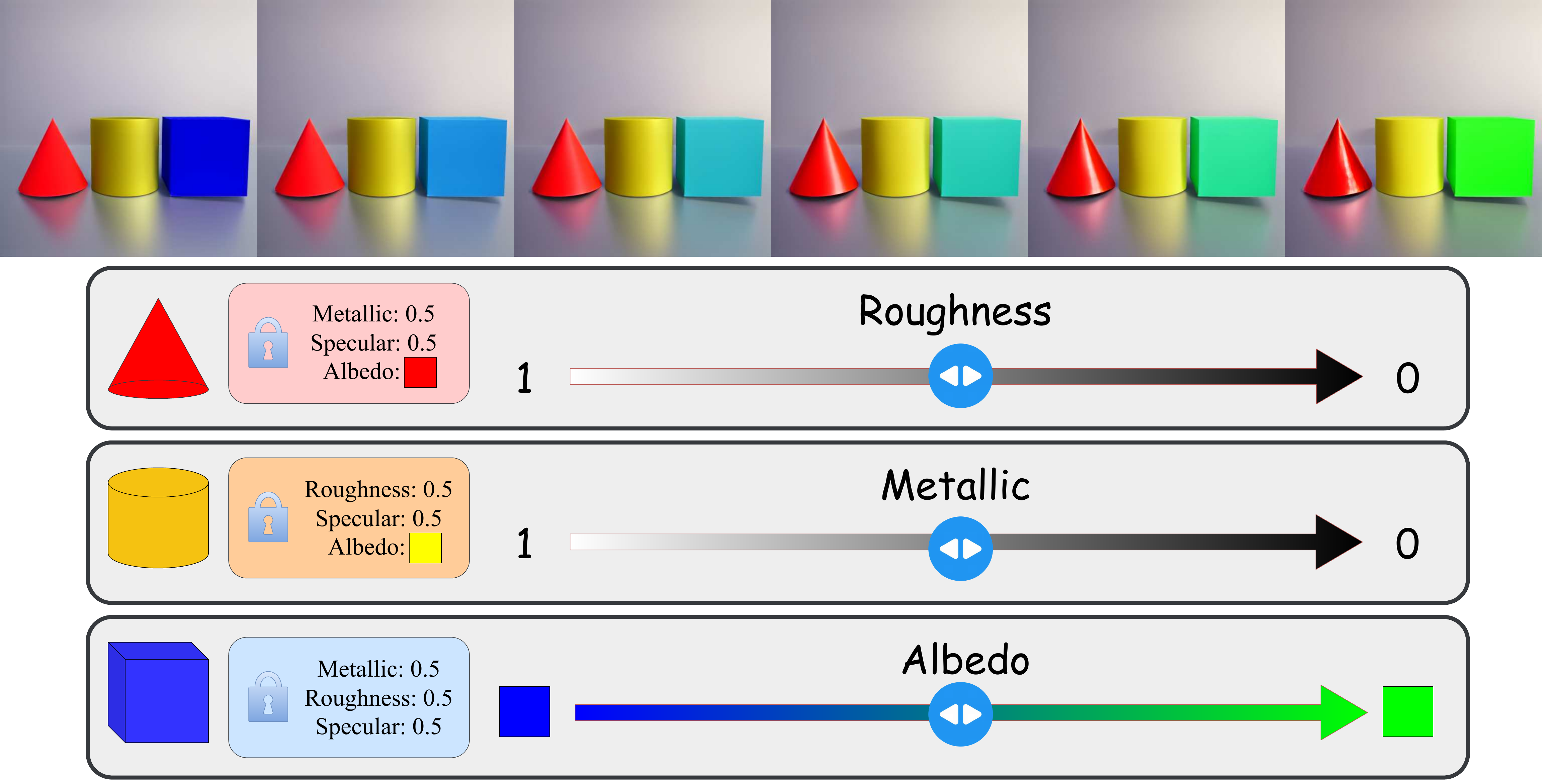}
    \caption{\textbf{Demonstration of controlled material editing.} Our model is
    capable of synthesizing smooth transitions for individual material
    properties while all other scene parameters are held constant. From left to
    right: the cone's roughness is interpolated from 1.0 to 0.0; the cylinder's
    metallic property from 0.0 to 1.0; and the cube's diffuse albedo from blue
    to green.}
    \label{fig:mat_edit}
\end{figure*}

\section{Implementation Details}
\subsection{Forward Rendering}
Our model is fine-tuned from the pre-trained Wan2.1 (1.3B) video diffusion model
\cite{wan2025}. Given that our task is conditioned on G-buffers rather than
text, we remove the original cross-attention layers from the transformer blocks.
We adapt the input embedder from VACE \cite{jiang2025vace}, modifying it to
process albedo interpolants instead of noisy inputs. The VACE context embedder
is repurposed to process both the G-buffer attribute tokens and the progressive
predictions from previous video chunks by expanding its input channel dimension.
For environment map conditioning, we employ a separate input embedder,
initialized with the same weights as the primary albedo embedder. Keyframe
features are processed frame-wise using dedicated 2D convolutional layers to
patchify the keyframe latents into spatially-aligned tokens.

The training process consists of two main stages. To manage the significant
memory consumption of the VAE decoder during pixel-space loss computation, we
train our model on short sequences of 5 frames. The initial training stage
consists of 20,000 steps at a 256$\times$256 resolution with a batch size of 32 and a
constant learning rate of \(5 \times 10^{-5}\). We then continue training for
another 20,000 steps at the full 512$\times$512 resolution, using a learning rate of
\(3 \times 10^{-5}\) with a 1,000-step warm-up period. For the final keyframe
guidance stage, the model is trained for an additional 20,000 iterations at
512$\times$512 with a learning rate of \(1 \times 10^{-4}\). All training and
experiments utilize the AdamW \cite{loshchilov2017decoupled} optimizer. Each
training stage takes approximately two days to complete on four NVIDIA A100
GPUs.

\subsection{Inverse Rendering}

\paragraph{Architecture Details.}
We adapt the pretrained \textit{RenderFlow} framework for inverse rendering by freezing the entire forward backbone—including the VAE encoder/decoder ($\mathcal{E}, \mathcal{D}$) and the DiT transformer blocks—and introducing a set of parameter-efficient adapter modules.
First, we replace the original G-buffer input projection with a trainable inverse embedder. This module maps the latent representation of the input RGB image, $z_{\text{rgb}} = \mathcal{E}(I_{\text{rgb}})$, into the token space required by the transformer. 
Second, to adapt the frozen self-attention mechanisms for decomposition tasks, we inject Low-Rank Adaptation (LoRA) modules into the query, key, and value projections of each block. 
Third, to control the target output modality (e.g., Albedo vs. Normal), we incorporate a prompt-based cross-attention layer after each self-attention block. Following RGB-X \cite{zeng2024rgb}, we use text embeddings of the target attribute as keys and values, allowing the model to switch tasks dynamically.
Finally, the transformer outputs are processed by lightweight, task-specific intrinsic heads (MLPs) that project the features into the appropriate latent space before decoding to further distinguish between modalities.

\paragraph{Training Objectives.}
We utilize the same bridge matching framework as the forward pass but apply it to the latent trajectory from the RGB input to the target intrinsic. We supervise the training using modality-specific reconstruction losses $\mathcal{L}_{\text{rec}}$ calculated in the pixel space:

\begin{enumerate}
    \item \textbf{Albedo}: We utilize a combination of $\mathcal{L}_1$ distance and a perceptual loss (LPIPS) to preserve both color accuracy and high-frequency details:
    \begin{equation}
        \mathcal{L}_{\text{albedo}} = \| \hat{I} - I_{\text{gt}} \|_1 + \lambda \mathcal{L}_{\text{LPIPS}}(\hat{I}, I_{\text{gt}})
    \end{equation}
    
    \item \textbf{Normal}: We employ a cosine similarity loss to enforce angular consistency between the predicted and ground truth normal vectors:
    \begin{equation}
        \mathcal{L}_{\text{normal}} = 1 - \frac{1}{N}\sum \frac{\langle \hat{I}, I_{\text{gt}} \rangle}{\| \hat{I} \|_2 \cdot \| I_{\text{gt}} \|_2}
    \end{equation}
    
    \item \textbf{Depth}: To handle scale ambiguity, we use the scale-and-shift invariant (SSI) loss~\cite{ranftl2020towards}:
    \begin{equation}
        \mathcal{L}_{\text{depth}} = \frac{1}{N} \sum \Delta_i^2 - \frac{1}{2N^2} \left( \sum \Delta_i \right)^2
    \end{equation}
    where $\Delta_i$ is the difference in log-depth at pixel $i$.
    
    \item \textbf{Material}: For roughness and metallic maps, we apply a standard $\mathcal{L}_1$ loss:
    \begin{equation}
        \mathcal{L}_{\text{material}} = \| \hat{I} - I_{\text{gt}} \|_1
    \end{equation}
\end{enumerate}

We train the adapter modules on the same dataset used for the forward model. The backbone weights remain frozen throughout the process. We use the AdamW optimizer with a learning rate of $1 \times 10^{-4}$ and a total batch size of 32. The model is trained for 20,000 iterations on 4 NVIDIA A100 GPUs. During training, we randomly sample a target modality and its corresponding text prompt for each batch to ensure the model learns to disentangle all intrinsics simultaneously.

\begin{figure}[ht]
    \centering
    \setlength{\tabcolsep}{0pt} %
    \renewcommand{\arraystretch}{0} %
    \begin{tabular}{cccc}
        \includegraphics[width=0.25\linewidth]{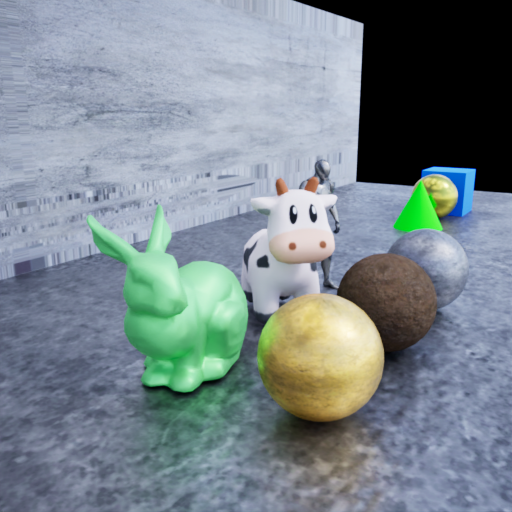} &
        \includegraphics[width=0.25\linewidth]{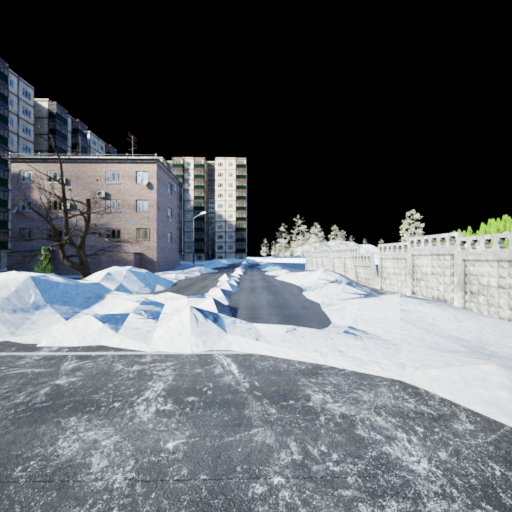} &
        \includegraphics[width=0.25\linewidth]{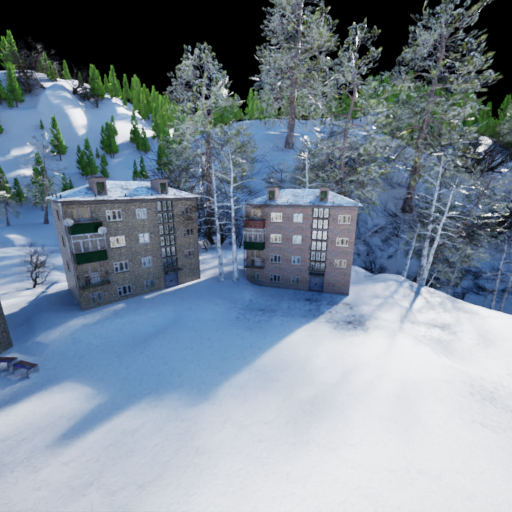} &
        \includegraphics[width=0.25\linewidth]{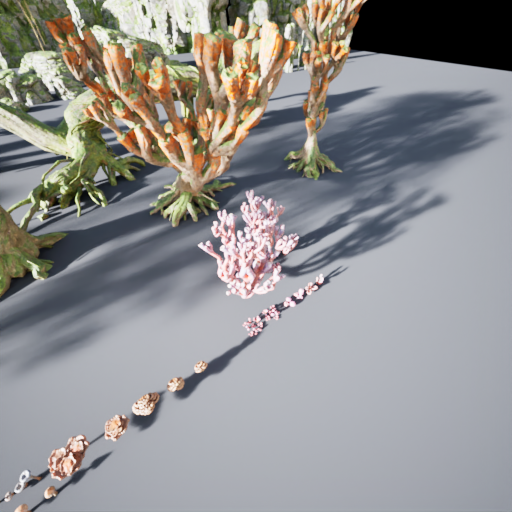}   \\
        \includegraphics[width=0.25\linewidth]{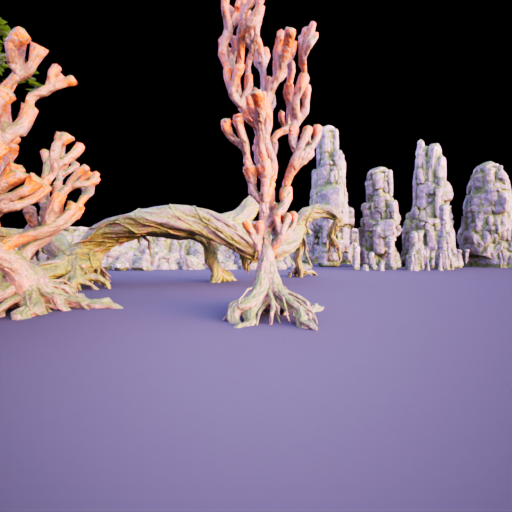} &
        \includegraphics[width=0.25\linewidth]{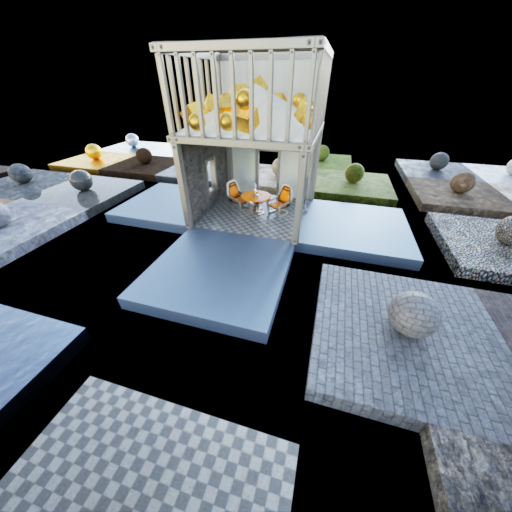} &
        \includegraphics[width=0.25\linewidth]{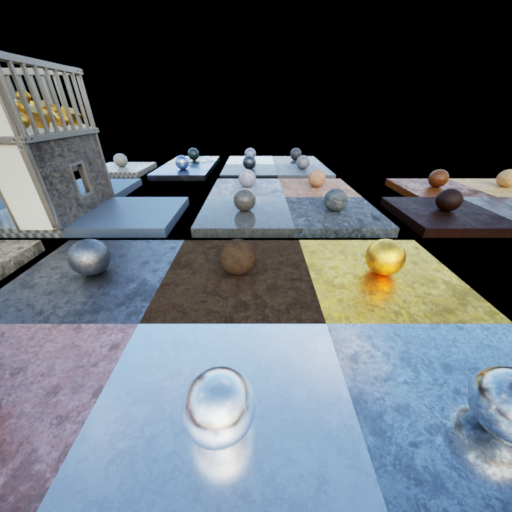} &
        \includegraphics[width=0.25\linewidth]{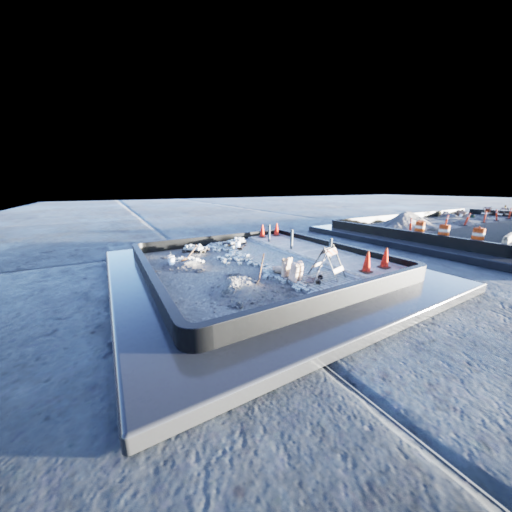}   \\
        \includegraphics[width=0.25\linewidth]{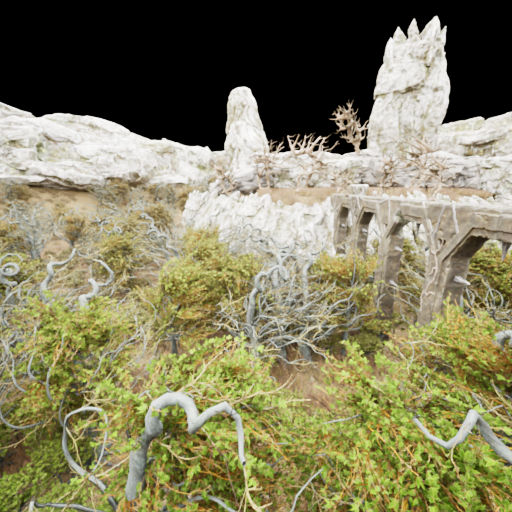} &
        \includegraphics[width=0.25\linewidth]{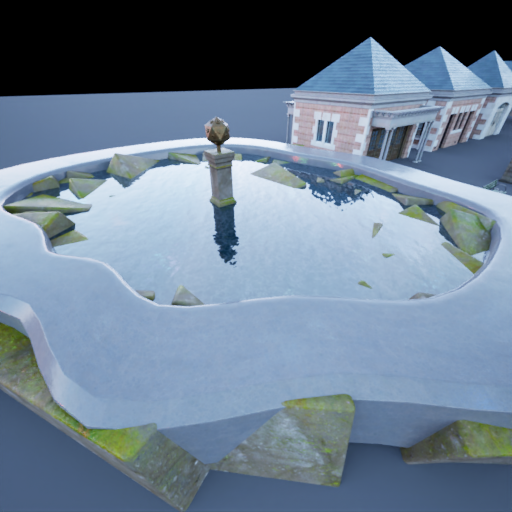} &
        \includegraphics[width=0.25\linewidth]{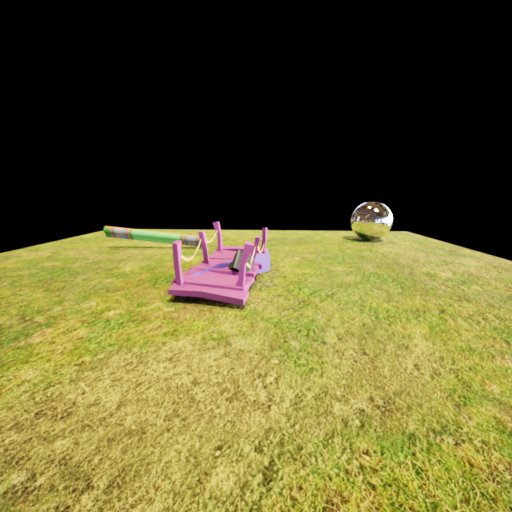} &
        \includegraphics[width=0.25\linewidth]{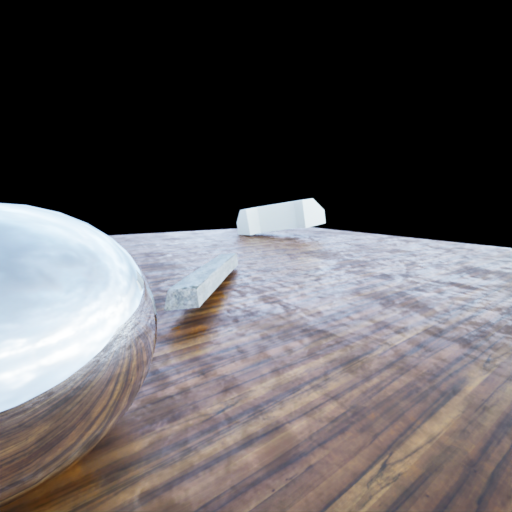}   \\
        \includegraphics[width=0.25\linewidth]{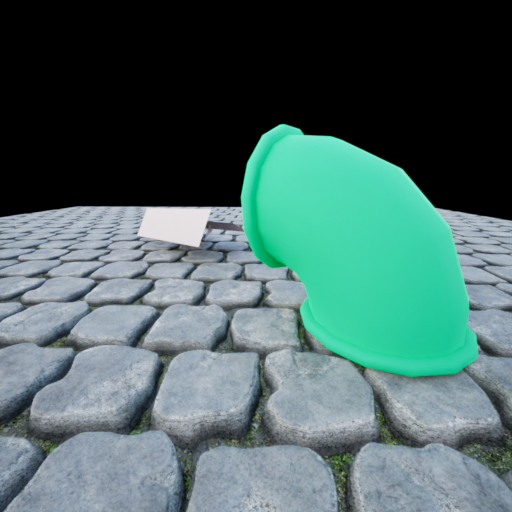} &
        \includegraphics[width=0.25\linewidth]{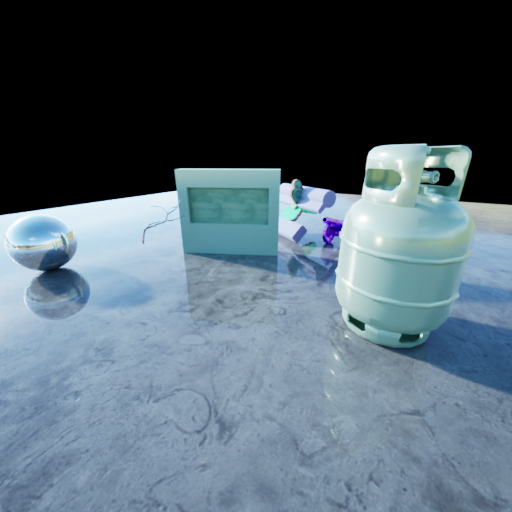} &
        \includegraphics[width=0.25\linewidth]{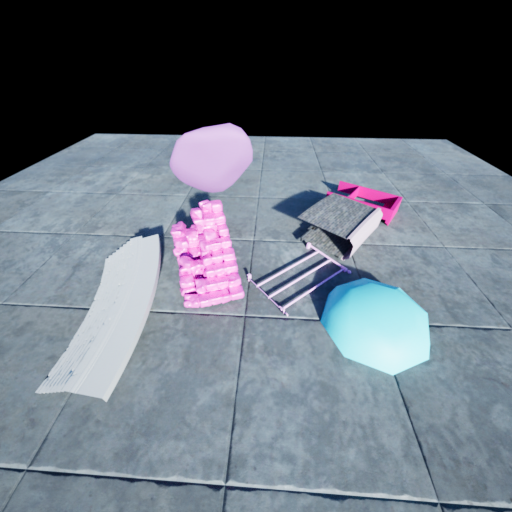} &
        \includegraphics[width=0.25\linewidth]{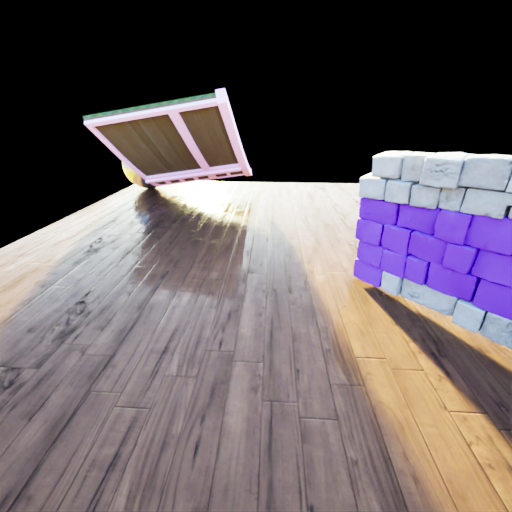}   \\
        \includegraphics[width=0.25\linewidth]{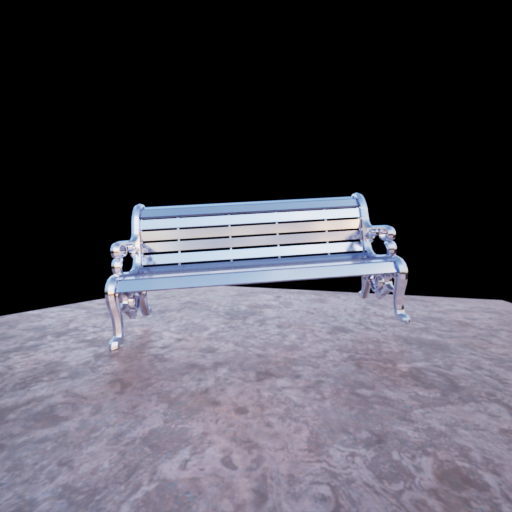} &
        \includegraphics[width=0.25\linewidth]{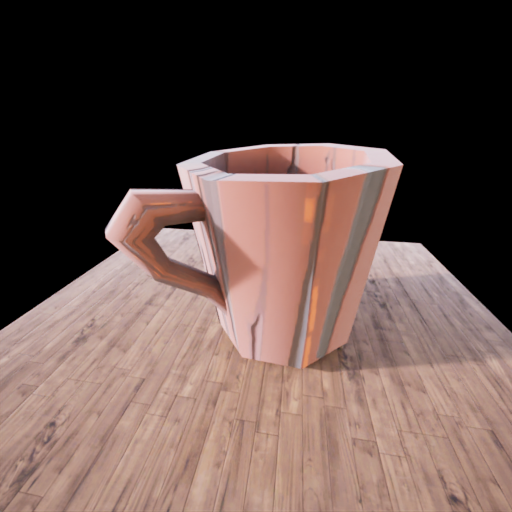} &
        \includegraphics[width=0.25\linewidth]{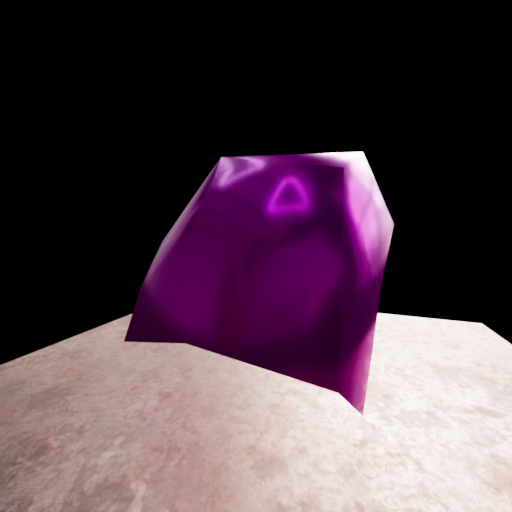} &
        \includegraphics[width=0.25\linewidth]{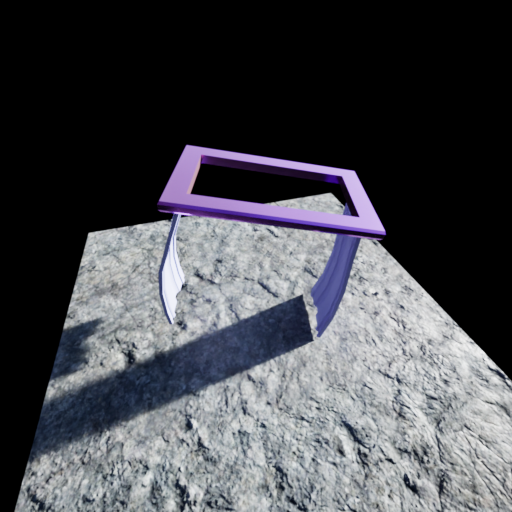}   \\
        \includegraphics[width=0.25\linewidth]{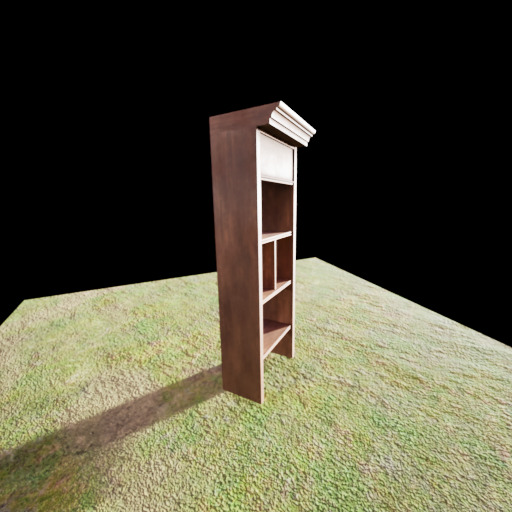} &
        \includegraphics[width=0.25\linewidth]{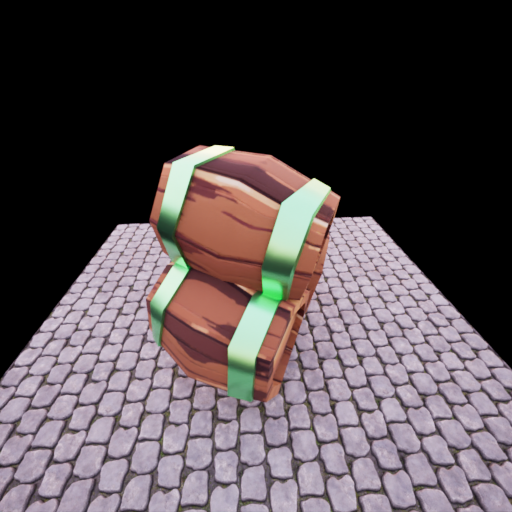} &
        \includegraphics[width=0.25\linewidth]{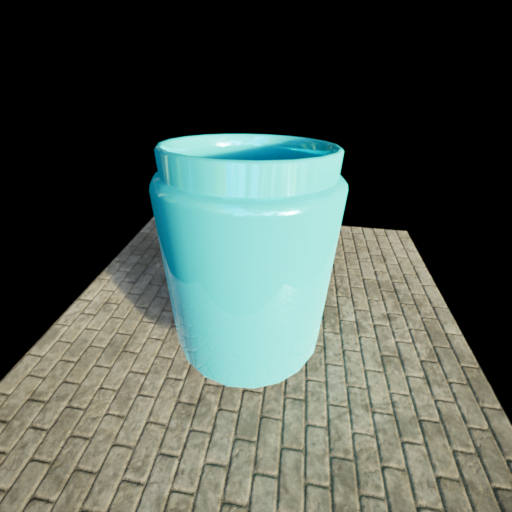} &
        \includegraphics[width=0.25\linewidth]{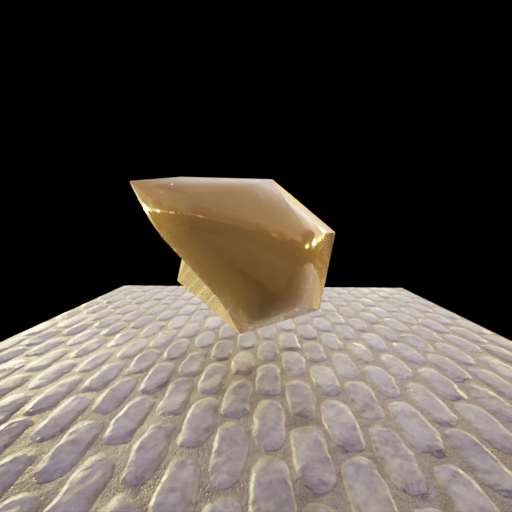} \\
    \end{tabular}
    \caption{A preview of our synthesized dataset, showcasing a variety of scenes, objects, and lighting conditions.}
    \label{fig:dataset_preview}
\end{figure}

\subsection{Input Buffer Selection}
We evaluate the choice of primary input buffer by comparing Albedo alone against a combined Albedo$\times$Irradiance input. Both variants are trained at low resolution for one day as a lightweight ablation. As shown in Table~\ref{tab:buffer_combination}, incorporating irradiance as a multiplicative prior improves PSNR by 1.9 dB, confirming that pre-multiplied lighting information provides a useful signal. However, to maintain a fair comparison with DiffusionRenderer~\cite{DiffusionRenderer}, which uses albedo as its primary input, we adopt the same protocol in our main experiments.

\begin{table}[ht]
    \centering
    \begin{tabular}{l|ccc}
        \toprule
        \textbf{Input} & PSNR $\uparrow$ & SSIM $\uparrow$ & LPIPS $\downarrow$ \\
        \midrule
        Albedo                    & 23.7 & 0.75 & 0.13 \\
        Albedo$\times$Irradiance  & 25.6 & 0.76 & 0.13 \\
        \bottomrule
    \end{tabular}
    \caption{\textbf{Buffer combination analysis.} Trained at low resolution for one day as a lightweight ablation. Combining albedo with irradiance as a multiplicative prior improves reconstruction quality.}
    \label{tab:buffer_combination}
\end{table}

\section{Evaluation of Keyframe Guidance}
We further evaluate the effectiveness of our keyframe guidance by analyzing its
performance under varying levels of keyframe sparsity. While the model was
trained with a fixed keyframe gap of 16 frames, the adapter's design allows for
flexible injection of arbitrary keyframes during inference. We conduct
experiments with keyframe gaps of 13, 17, 25, and 49 frames, and include a
baseline without keyframe guidance. Table \ref{tab:keyframe_gap} summarizes the
quantitative results. The experiment shows that, consistent with the intuition,
increasing the keyframe gap leads to a degradation in reconstruction quality.
This confirms that more frequent keyframes provide more effective guidance.
Notably, even with a large keyframe gap of 49 frames, the model with keyframe
guidance still significantly outperforms the unguided baseline, demonstrating
the robustness of the guidance mechanism. The consistency achieved by the
progressive inference is shown in the \textbf{Supplementary Videos}.

\begin{table}[ht]
    \centering
    \setlength{\tabcolsep}{1pt}
    \begin{tabular}{l|ccc}
        \toprule
        \textbf{Method} & PSNR $\uparrow$   & SSIM $\uparrow$  & LPIPS $\downarrow$ \\
        \midrule
        w/o keyframes   & 24.022            & 0.899            & 0.112              \\
        \midrule
        13 Gap          & \textbf{29.716}(29.427) & \textbf{0.920}(0.918) & \textbf{0.089}(0.091)     \\
        17 Gap          & 29.071(28.812)            & 0.917(0.916)            & 0.092(0.094)             \\
        25 Gap          & 26.574(26.507)            & 0.913(0.911)           & 0.098(0.099)              \\
        49 Gap          & 25.917(25.737)            & 0.909(0.909)           & 0.102(0.104)           \\
        \bottomrule
    \end{tabular}
    \caption{\textbf{Quantitative evaluation of the impact of keyframe gaps.}
        Reconstruction quality degrades as the distance between keyframes increases.
        Values in parentheses report metrics calculated exclusively on non-keyframe
        frames to evaluate guidance performance.}
\label{tab:keyframe_gap}
\end{table}

\section{Demonstration of Material Editing}
In this section, we evaluate our model's ability to render materials in a
controlled and disentangled manner. We conducted a synthetic experiment for
plausible material editing. We generated image sequences where specific material
properties were dynamically varied for individual objects, while all other scene
parameters, such as camera path and environment lighting, remained fixed.

The scene, shown in Figure \ref{fig:mat_edit}, contains three primitive objects:
a red cone, a yellow cylinder, and a blue cube. Throughout the sequence, we
linearly interpolate a single material parameter for each object while holding
its other properties constant:
\begin{itemize}
    \item \textbf{Cone (Roughness):} The cone's diffuse albedo is fixed to red,
    with metallic and specular values all set to 0.5. Its roughness parameter is
    linearly interpolated from 1.0 (fully diffuse) to 0.0 (perfectly smooth).
    \item \textbf{Cylinder (Metallic):} The cylinder's roughness and specular
    \item \textbf{Cone (Roughness):} The cone's diffuse albedo is fixed to red,
    values are fixed at 0.5 and 0.5. Its metallic parameter is linearly
    interpolated from 0.0 (dielectric) to 1.0 (fully metallic).
    \item \textbf{Cube (Albedo):} The cube's roughness and metallic values are
    fixed at 0.5 and 0.5. Its diffuse albedo is linearly interpolated in RGB
    space from blue (0,0,1) to green (0,1,0).
\end{itemize}
The results illustrate that our model can faithfully render the smooth
transitions between material properties, synthesizing the changing appearance
from diffuse to specular reflections on the cone, dielectric to conductive
properties on the cylinder, and the color shift on the cube. This highlights the
model's potential for downstream applications, such as interactive material editing.

\section{Analysis on inference steps}
Our model is trained in a 4-step bridge matching framework, making it possible
to either choose to perform single-step or multi-step (2 or 4 steps) inference.
In our experiments, we empirically found that single-step inference consistently
outperforms multi-step inference across all quantitative metrics. We attribute
this to error propagation: our model is trained to predict the final rendered
latent \(\hat{\mathbf{z}}_1\) directly from the initial albedo latent
\(\mathbf{z}_0\). Due to the inaccuracy of each network evaluation, intermediate
errors will be accumulated, resulting in greater offsets from the ground truth.

\begin{figure}[t]
    \centering
    \includegraphics[width=\linewidth]{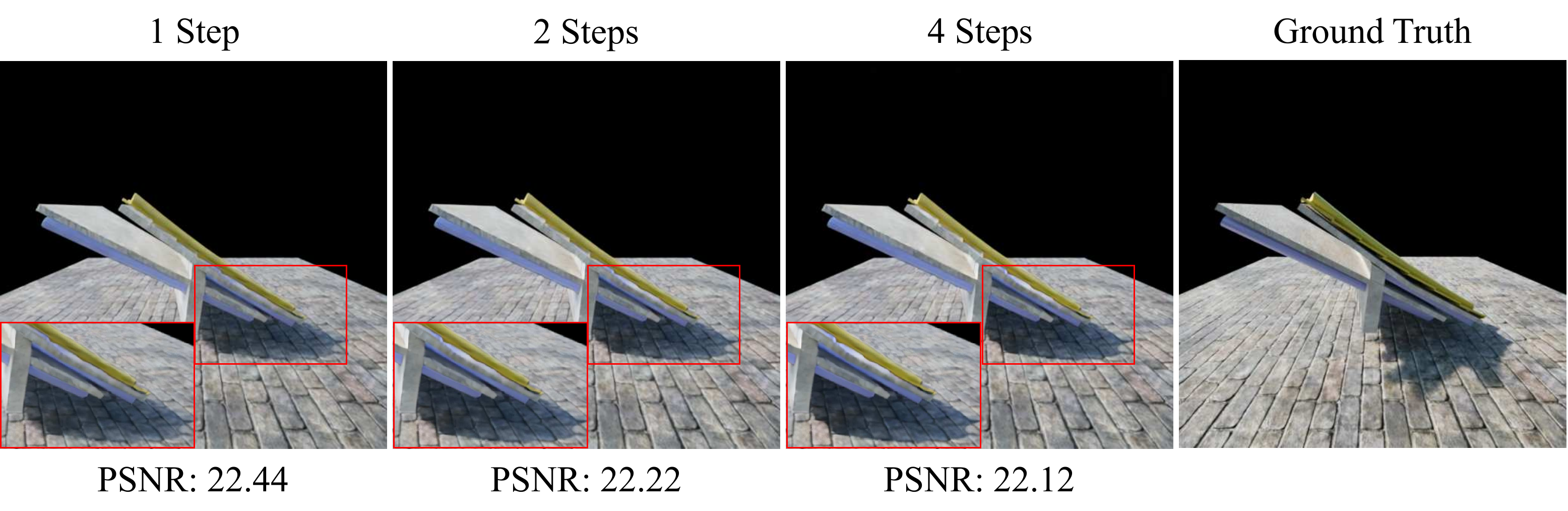}
    \vspace{4pt}
    \includegraphics[width=\linewidth]{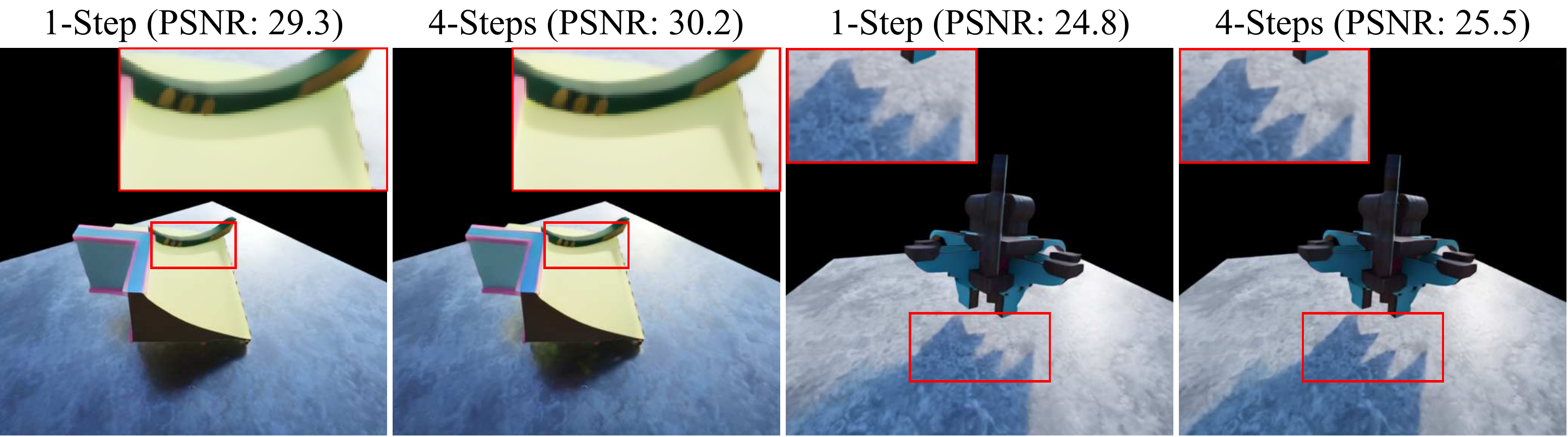}
    \caption{\textbf{Analysis on inference steps.} (a) Top: Multi-step inference
    can generate sharper shadows but leads to lower metrics due to misalignments
    with ground truth. (b) Bottom: In favorable cases where generated content
    aligns with ground truth, 4-step inference yields sharper and more detailed
    results.}
    \label{fig:inference_steps}
\end{figure}

Qualitatively, as shown in Figure \ref{fig:inference_steps}, we observe that
multi-step inference can produce visually sharper high-frequency details, such
as contact shadows. We hypothesize that this apparent sharpness comes at the
cost of geometric and photometric accuracy. The iterative process may amplify
certain features, making them visually prominent but causing them to deviate
from the ground truth. Therefore, while multi-step inference can create
plausibly sharp details, the single-step approach achieves a better balance of
visual quality and quantitative accuracy. Nevertheless, in cases where the
generated content aligns well with the ground truth, 4-step inference can
produce better results both visually and quantitatively.
Figure~\ref{fig:inference_steps}(b) illustrates such a case, where 4-step
inference yields sharper and more detailed outputs compared to single-step
inference.

\section{Dataset Curation}
Our dataset was synthesized using the \textit{Movie Render Queue} in Unreal
Engine 5. The dataset is composed of two main sources: pre-existing,
artist-crafted environments and procedurally generated synthetic scenes. The
artist-crafted environments were sourced from the Unreal Engine Marketplace
\cite{ue_marketplace} and include professionally built scenes with high-quality
assets. To adapt these scenes for our task, we removed all explicit light
sources, relying solely on image-based lighting from HDR environment maps. We
also manually modified materials to eliminate transparent and translucent
surfaces, ensuring that the G-buffers provide a complete and unambiguous
representation of the scene geometry.

Rendering these complex, pre-existing environments is time-intensive, often
taking over one hour for a 100-frame sequence. To significantly increase data
diversity and volume, we developed a procedural pipeline to generate synthetic
scenes. This pipeline randomly places objects from a curated asset collection
onto a ground plane and assigns them dynamic materials with randomized
properties. This automated approach reduced rendering time to approximately 15
minutes per 100-frame sequence. For the artist-crafted environments, we defined
fixed camera trajectories, whereas for the synthetic scenes, we primarily
utilized orbital camera paths. To further diversify lighting conditions, we
augmented the environment maps from the pre-existing scenes with additional
HDRIs from Poly Haven \cite{poly_haven}.

In total, our asset library contains over 4,000 unique meshes and more than 30
HDR environment maps. The final dataset consists of approximately 30,000 frames
from artist-crafted environments and 100,000 frames from synthetic scenes. All
frames were rendered at a resolution of 512x512 pixels with 256 samples per
pixel (SPP) and subsequently denoised using Intel Open Image Denoise
\cite{OpenImageDenoise}. While our dataset is smaller than that used by
DiffusionRenderer \cite{DiffusionRenderer}, which contains 150,000 videos, our
model's efficiency and training strategy allow it to achieve strong performance.
A preview of our dataset is provided in Figure \ref{fig:dataset_preview}.

\section{Limitations and Future Work.} 
\begin{figure}[t]
       \centering
       \includegraphics[width=\columnwidth]{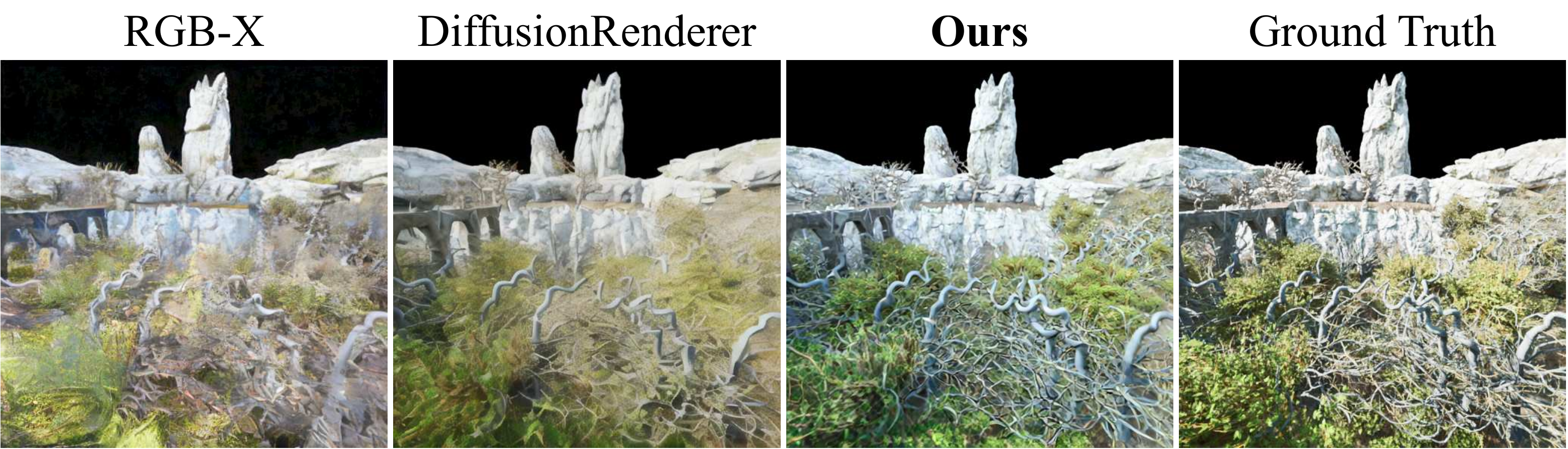} \\
       \includegraphics[width=\columnwidth]{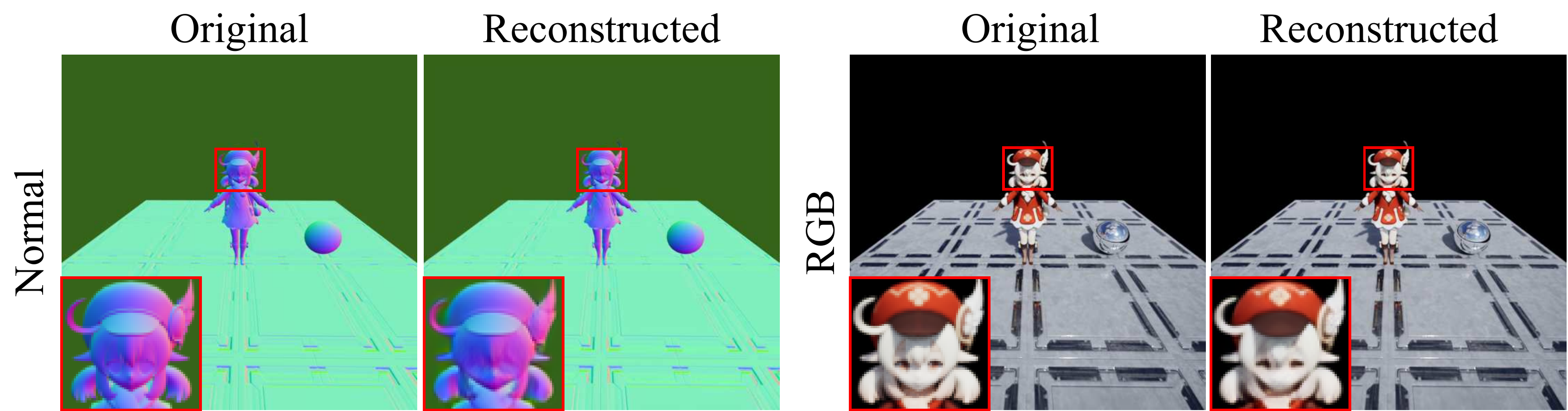}
       \caption{\textbf{Failed cases and analysis.} Top: Rendered images for scenes with
              highly complex geometries. Bottom: The VAE compression leads to loss of
              fine-grained details}
       \label{fig:failure_cases}
\end{figure}

\begin{figure}[t]
       \centering
       \includegraphics[width=\columnwidth]{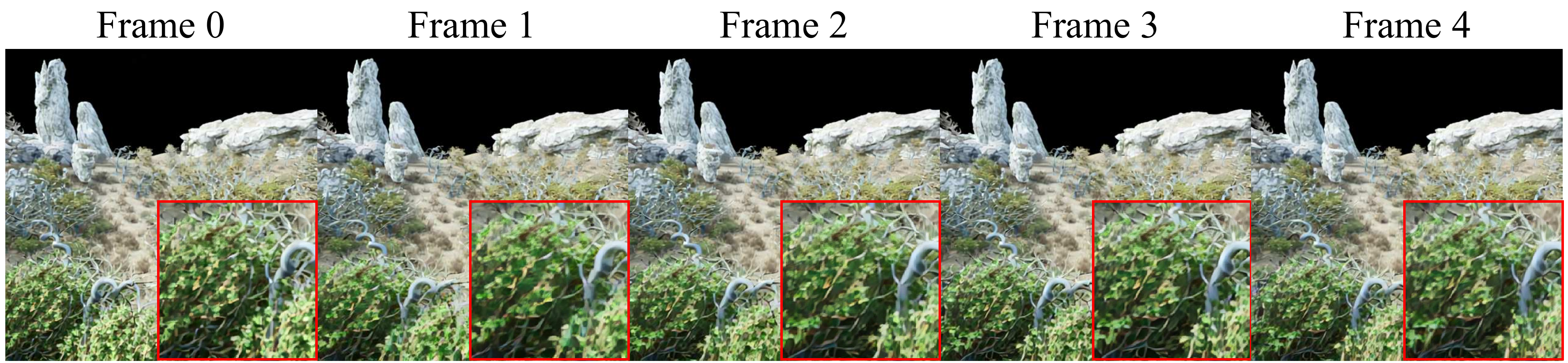}
        \caption{\textbf{Analysis on video latent downsampling.} The causal design of the VAE encoder leads to the phenomenon that the initial frame retains sharpness while subsequent frames exhibit progressive blurring due to temporal downsampling dependencies.}
       \label{fig:video_compress}
\end{figure}

While RenderFlow demonstrates significant
advancements, several avenues for improvement remain. First, the generalization
of our model is limited by the diversity of our current dataset. Expanding the training data to include a broader
range of lighting phenomena and geometric complexity would enhance the model's
robustness. Second, the VAE's latent space can cause information loss,
potentially limiting the reconstruction of details in highly
complex geometries as shown in Figure \ref{fig:failure_cases}. Furthermore, the causal convolution within the Wan VAE 
encoder performs temporal compression that affects frame consistency. Specifically, later frames tend to 
exhibit progressive blurring compared to the sharp initial frame due to causal dependencies in the latent 
space as shown in Figure \ref{fig:video_compress}. Finally, the VAE
encoder and decoder still constitute a significant computational bottleneck.
Future work could explore more efficient encoding strategies for G-buffers
to enhance quality and speed. We believe that with larger and more diverse
datasets, more efficient model architectures, our proposed paradigm has the potential to
achieve greater performance in neural rendering and video synthesis.

\section{Additional Qualitative Results}
\begin{figure*}[b]
    \centering
    \includegraphics[width=\textwidth]{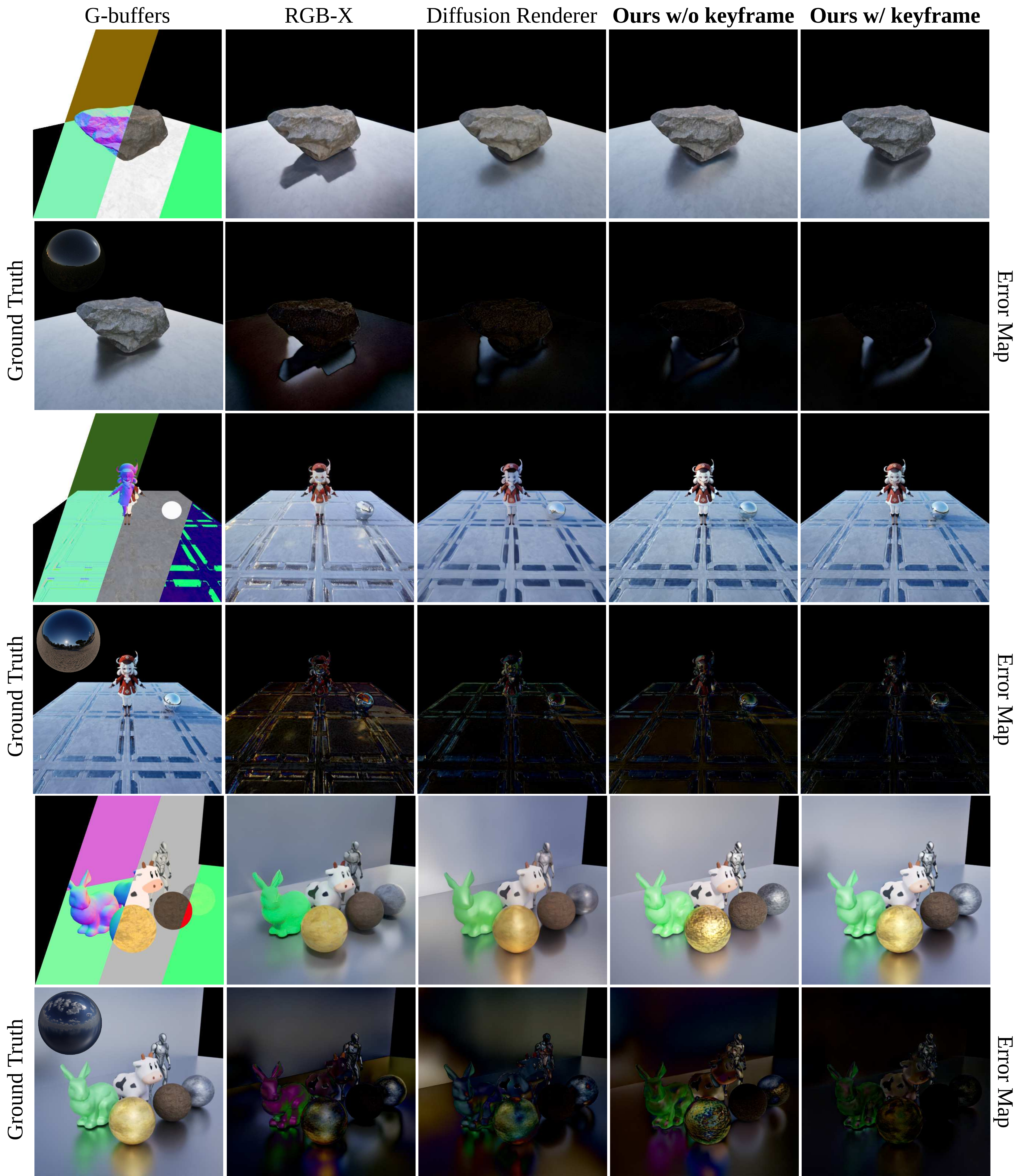}
    \caption{\textbf{Additional qualitative forward rendering results.} We provide more
        qualitative comparisons with error maps to highlight the differences
        between our predictions and the ground truth. For keyframe guidance, we 
        sample keyframes with a 25-frame gap. Our method clearly outperforms the 
        baseline methods. In addition, with the keyframe guidance our model achieves
        the best results among all models.}
    \label{fig:more_results1}
\end{figure*}

\begin{figure*}[b]
    \centering
    \includegraphics[width=\textwidth]{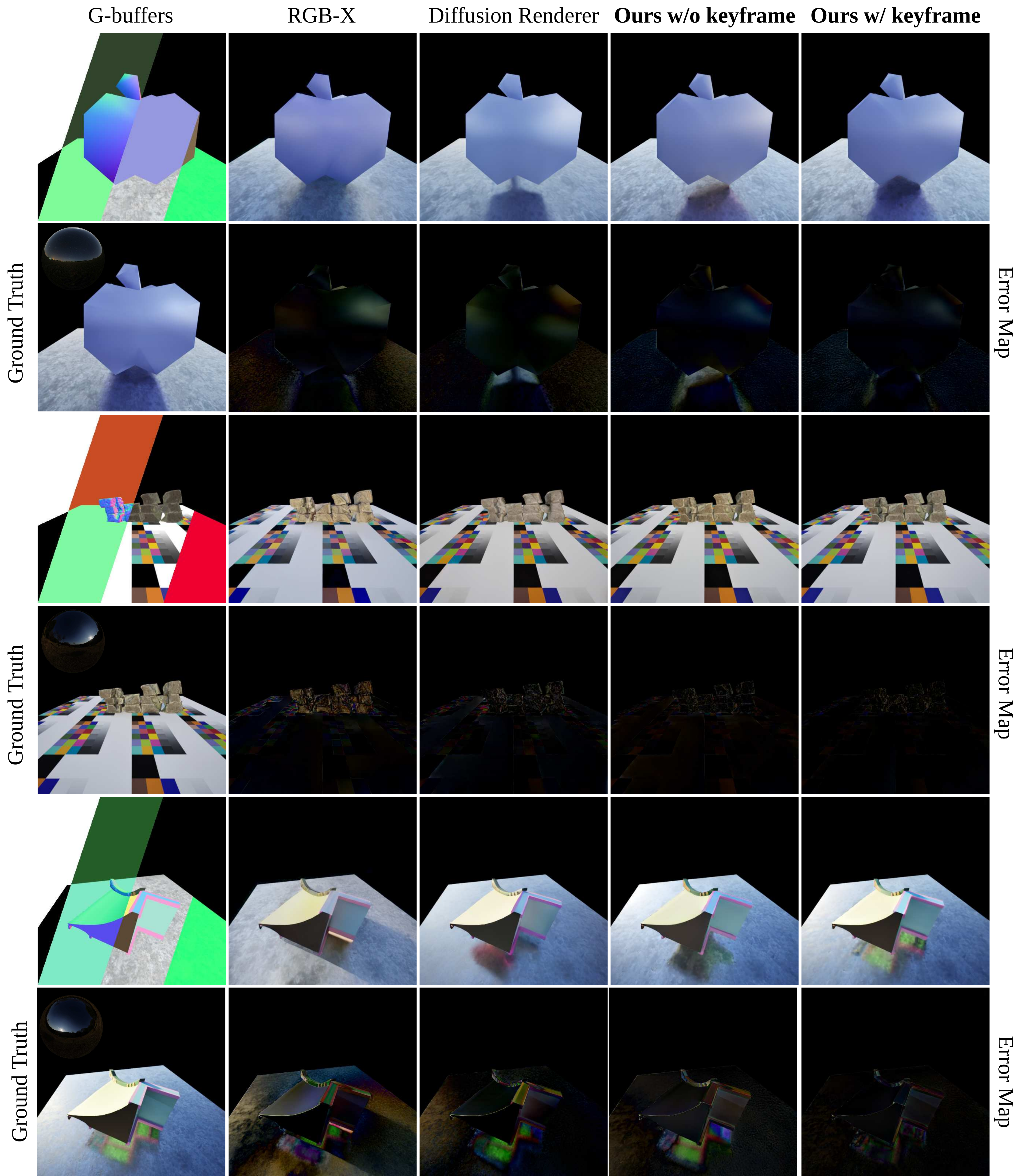}
    \caption{\textbf{Additional qualitative forward rendering results (continued).}}
    \label{fig:more_results2}
\end{figure*}

We provide additional qualitative comparisons in Figures \ref{fig:more_results1}
and \ref{fig:more_results2}, including error maps to highlight differences
between the ground truth and our predictions. The results show that our model
achieves performance comparable to DiffusionRenderer in synthesizing complex
effects like shadows and detailed reflections, while better preserving the
underlying geometry of the scene. Resolving misalignments in shadows and
reflections is challenging, as the information provided by G-buffers alone can
be ambiguous. However, we demonstrate that with keyframe guidance, our model can
leverage information from the reference view to generate these high-frequency
details with greater physical plausibility and accuracy.

\begin{figure*}[b]
    \centering
    \includegraphics[width=\textwidth]{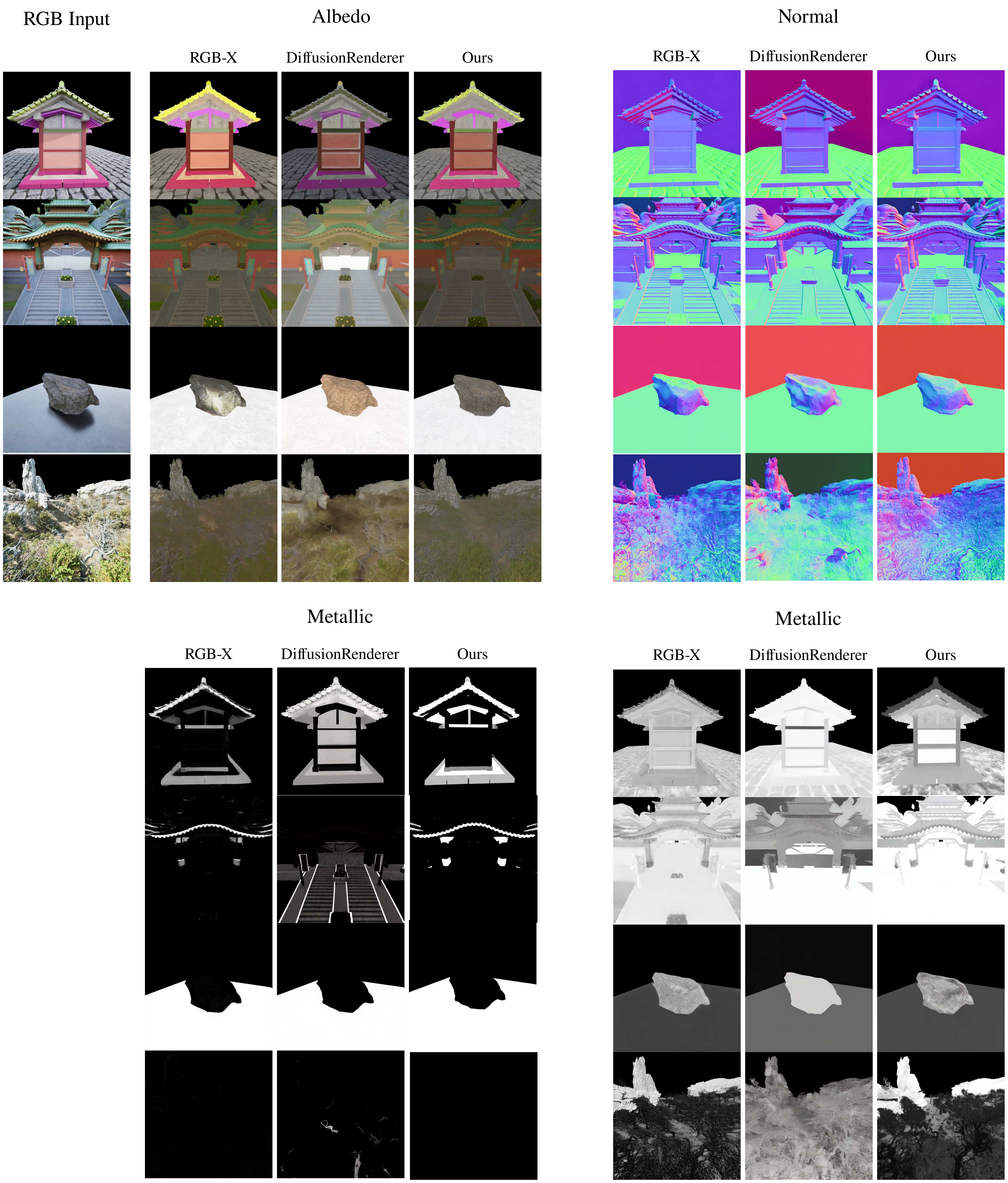}
    \caption{\textbf{Additional qualitative inverse rendering results.}}
    \label{fig:more_results3}
\end{figure*}

\end{document}